\documentclass[11pt]{article}

\usepackage[preprint]{acl}

\usepackage{times}
\usepackage{latexsym}
\usepackage[T1]{fontenc}
\usepackage[utf8]{inputenc}
\usepackage{microtype}
\usepackage{inconsolata}

\usepackage{graphicx}
\usepackage{booktabs}
\usepackage{multirow}
\usepackage{amsmath}
\usepackage{amssymb}
\usepackage{url}
\usepackage{enumitem}
\usepackage{siunitx}
\usepackage{float}
\usepackage{adjustbox}
\usepackage{algorithm}
\usepackage{algpseudocode}
\usepackage{listings}
\usepackage{tcolorbox}
\tcbuselibrary{breakable,skins,listings}

\usepackage{caption}
\usepackage{titlesec}
\titlespacing*{\section}{0pt}{1.0ex plus 0.3ex minus 0.2ex}{0.4ex plus 0.2ex}
\titlespacing*{\subsection}{0pt}{0.8ex plus 0.3ex minus 0.2ex}{0.3ex plus 0.1ex}
\titlespacing*{\paragraph}{0pt}{0.4ex plus 0.2ex minus 0.1ex}{0.5em}

\usepackage{xcolor}

\usepackage{tikz}
\usetikzlibrary{positioning, arrows.meta, shapes.geometric, calc, fit, backgrounds, decorations.pathreplacing, shadows.blur}
\definecolor{ThesisLine}{HTML}{D0D8E3}
\definecolor{ThesisNavy}{HTML}{2C3E50}
\definecolor{PromptOutputBg}{HTML}{FAF6EE}
\definecolor{PromptOutputFrame}{HTML}{B89B6A}
\definecolor{PromptOutputTitle}{HTML}{8A7040}

\tcbset{
  promptbox/.style={
    enhanced, colback=gray!3, colframe=gray!60, boxrule=0.5pt, arc=2pt,
    left=6pt, right=6pt, top=6pt, bottom=6pt, fontupper=\small\ttfamily,
    before upper={\parindent0pt\parskip4pt}
  },
  systemprompt/.style={
    enhanced, breakable, colback=green!3, colframe=green!40!black,
    coltitle=black, fonttitle=\small\bfseries, boxrule=0.5pt, arc=2pt,
    left=6pt, right=6pt, top=6pt, bottom=6pt, fontupper=\small\ttfamily,
    before upper={\parindent0pt\parskip4pt},
    attach boxed title to top left={yshift=-2mm, xshift=4mm},
    boxed title style={colback=green!15, colframe=green!40!black, arc=1pt, boxrule=0.3pt}
  },
  humanprompt/.style={
    enhanced, breakable, colback=blue!2, colframe=blue!40,
    coltitle=black, fonttitle=\small\bfseries, boxrule=0.5pt, arc=2pt,
    left=6pt, right=6pt, top=6pt, bottom=6pt, fontupper=\small\ttfamily,
    before upper={\parindent0pt\parskip4pt},
    attach boxed title to top left={yshift=-2mm, xshift=4mm},
    boxed title style={colback=blue!10, colframe=blue!40, arc=1pt, boxrule=0.3pt}
  },
  codebox/.style={
    enhanced, breakable, colback=gray!3, colframe=gray!50, coltitle=black,
    fonttitle=\small\bfseries, boxrule=0.5pt, arc=2pt,
    left=4pt, right=4pt, top=4pt, bottom=4pt,
    fontupper=\scriptsize\ttfamily,
    before upper={\parindent0pt\parskip2pt},
    attach boxed title to top left={yshift=-2mm, xshift=4mm},
    boxed title style={colback=gray!15, colframe=gray!50, arc=1pt, boxrule=0.3pt}
  }
}

\newtcolorbox{promptbox}[1][]{enhanced, colback=gray!3, colframe=gray!60, boxrule=0.5pt,
  arc=2pt, left=6pt, right=6pt, top=6pt, bottom=6pt,
  fontupper=\small\ttfamily, before upper={\parindent0pt\parskip4pt}, breakable, #1}

\lstdefinestyle{pythonstyle}{
  language=Python, basicstyle=\scriptsize\ttfamily,
  keywordstyle=\color{blue!70!black}, commentstyle=\color{green!50!black},
  stringstyle=\color{orange!70!black}, showstringspaces=false,
  breaklines=true, frame=none, numbers=none, aboveskip=2pt, belowskip=2pt
}

\setlist[itemize]{leftmargin=*, nosep}
\setlist[enumerate]{leftmargin=*, nosep}

\title{Quality Without Usefulness: LLM-Generated XAI Narratives as Trust Heuristics Rather Than Decision Aids}

\author{
    \textbf{Fabian Lukassen}$^{1}$,
    \textbf{Jan Herrmann}$^{2}$,
    \textbf{Christoph Weisser}$^{3}$,\\
    \textbf{Alexander Silbersdorff}$^{1}$,
    \textbf{Benjamin Saefken}$^{4}$,
    \textbf{Thomas Kneib}$^{1}$\\
    $^{1}$University of G\"ottingen \quad $^{2}$BASF SE \quad $^{3}$Hochschule Bielefeld \quad $^{4}$TU Clausthal \\
    \texttt{fabian.lukassen@stud.uni-goettingen.de}, \texttt{jan.herrmann@basf.com},\\
    \texttt{christoph.weisser@hsbi.de}, \texttt{asilbersdorff@uni-goettingen.de},\\
    \texttt{benjamin.saefken@tu-clausthal.de}, \texttt{tkneib@uni-goettingen.de}
}

\begin{document}
\maketitle
\begin{abstract}
Prior work shows that Large Language Models (LLMs) can transform Explainable AI (XAI) outputs into Natural Language Explanations (NLEs) that score highly on quality metrics such as plausibility, coherence, and comprehensibility. But does explanation \textit{quality} translate to practical \textit{usefulness}? We investigate this question in a time-series energy forecasting domain through five controlled experiments (2{,}730 judgments across 60 test instances), each operationalising a distinct facet of usefulness studied in the XAI literature. Holding NLE quality constant at the high levels established by a prior factorial study, we find that NLEs do not improve task accuracy on any of the five tasks, while inflating self-reported confidence. A placebic control shows that this confidence boost is driven by text \textit{presence} rather than content. In an out-of-distribution detection task, NLEs reduce the LLM judge's ability to flag unreliable predictions, providing false reassurance that masks model failure. We characterise these findings as the \textbf{Quality-Usefulness Gap} and argue that evaluation of the XAI-to-NLE pipeline must extend beyond text-quality metrics to downstream task performance.\footnote{\href{https://github.com/fabian-lu/quality-usefulness-gap}{Code and data URL}.}
\end{abstract}

\section{Introduction}
\label{sec:intro}
Post-hoc XAI methods such as SHAP \citep{lundberg2017unified} produce feature-attribution outputs that presuppose statistical literacy the intended audiences -- domain experts, decision makers, regulators -- typically lack \citep{arrieta2020explainable, miller2019explanation}. Natural Language Explanations (NLEs) generated by Large Language Models (LLMs) translate those outputs into prose accessible to non-experts. A growing body of work now builds such NLE-from-XAI pipelines across tabular \citep{martens2024xaistories, zytek2024explingo, Dwiyanti2025ContextualSHAP, swamy2025illuminate}, graph \citep{cedro2025graphxain}, and time-series \citep{aksu2024xforecast} settings, all reporting consistently high quality scores. A recent factorial study \citep{fabian2025fromxaitostories} confirms this at scale -- across 4~ML models, 3~XAI conditions, 3~LLMs, and 8~prompting strategies (660~NLEs), G-Eval scores reach 4.0--4.8/5. But quality there was measured on text-artefact properties only: plausibility, coherence, comprehensibility.

\begin{figure}[t]
\centering
\includegraphics[width=\columnwidth]{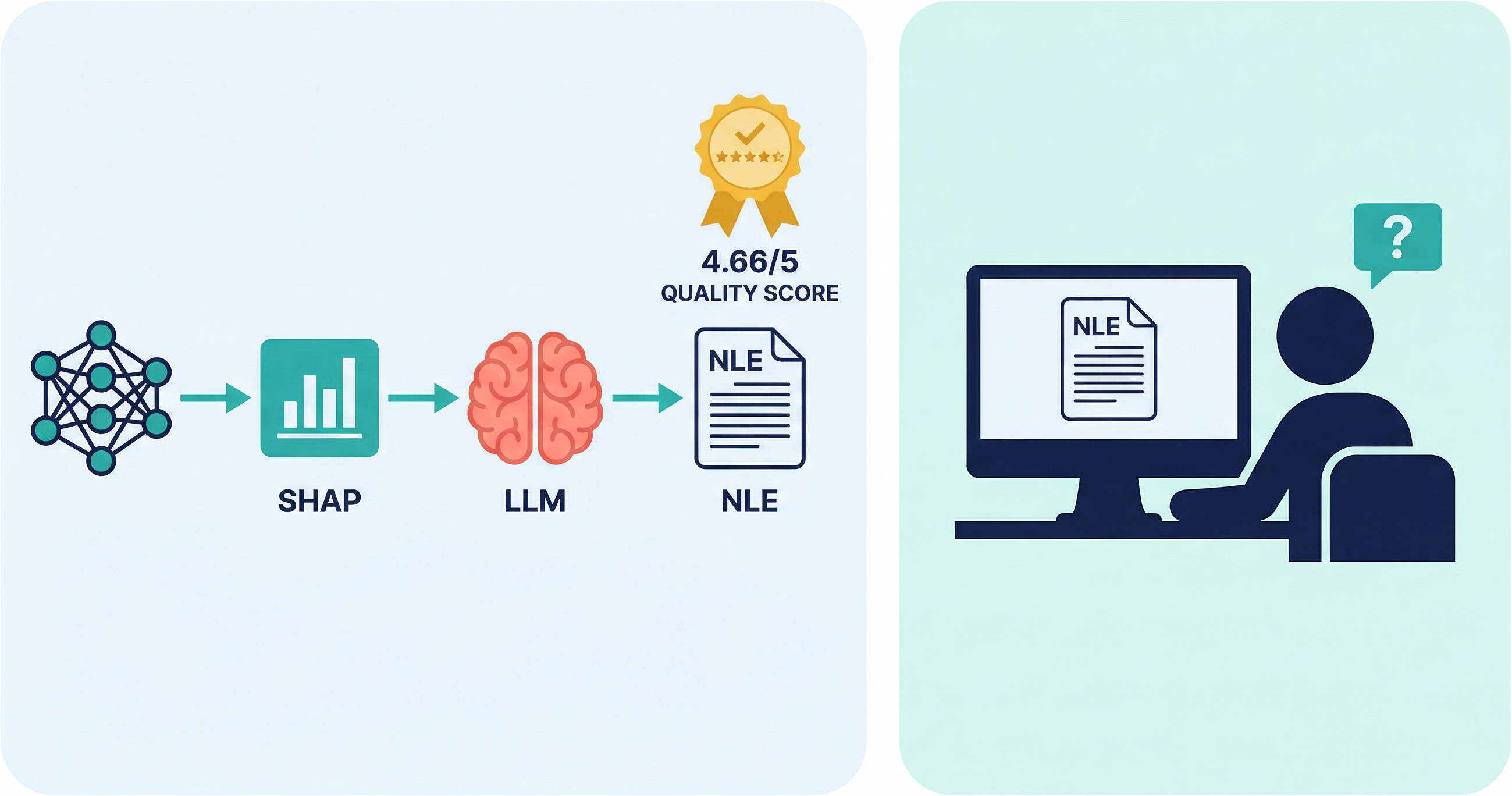}
\caption{An LLM narrates a prediction and XAI attributions into a high-quality NLE. We test whether such NLEs help downstream decisions.}
\label{fig:hook}
\end{figure}

Quality, however, is not utility. Research on the visual XAI outputs that these NLE pipelines build on -- feature-importance charts, saliency maps, attribution plots -- has repeatedly shown that they often fail to improve, and sometimes harm, downstream decision-making \citep{Bansal2021DoesTW, Jesus2021HowCI, schemmer2022should, bucinca2021trust}. That work draws a distinction between explanation \emph{quality} (text-artefact properties; \citealp{Nauta2023SystematicReview}) and \emph{usefulness} (measurable impact on downstream decisions -- simulatability, task performance, trust calibration; \citealp{doshi2017towards, jacovi2021formalizing}). This distinction has not been carried over to LLM-generated NLEs.

We test it. Holding NLE quality constant at the levels established by \citet{fabian2025fromxaitostories}, we run five controlled experiments, each covering a distinct facet of usefulness studied in the XAI literature, on a household energy forecasting task. Across 2{,}730 LLM-judge judgments, NLEs fail to improve accuracy on any of the five tasks; in the most consequential setting -- detecting out-of-distribution inputs -- they appear to \emph{reduce} the judge's ability to flag unreliable predictions. We characterise this pattern as the \textbf{Quality-Usefulness Gap}.

\paragraph{Contributions:} (1)~Empirical evidence for the Quality-Usefulness Gap in LLM-generated NLEs from XAI outputs. (2)~Diagnosis of two false-reassurance mechanisms behind that pattern: \emph{confidence inflation} driven by text presence rather than content (Experiments E1, E2), and \emph{rationalisation} of anomalous inputs (E5). (3) Together, these motivate a broader argument that NLE evaluation must move beyond quality metrics to task-based usefulness measures, just as standard XAI evaluation has done.

\section{Related Work}
\label{sec:related}

\paragraph{Post-hoc XAI.} SHAP \citep{lundberg2017unified}, the attribution method used throughout this paper, grounds per-feature importances in Shapley values from cooperative game theory: for prediction $f(x)$ with feature set $\mathcal{F}$, each feature receives
\begin{equation}
\phi_i = \sum_{S \subseteq \mathcal{F} \setminus \{i\}} \tfrac{|S|!(|\mathcal{F}|-|S|-1)!}{|\mathcal{F}|!}\, [f(x_{S \cup \{i\}}) - f(x_S)],
\end{equation}
satisfying $f(x) = \phi_0 + \sum_i \phi_i$; TreeSHAP \citep{lundberg2020local} computes this exactly in polynomial time for tree models. SHAP is widely adopted \citep{guidotti2018survey} but requires statistical fluency to interpret, particularly in time-series settings with autoregressive lags \citep{theissler2022explainable, zytek2024llms}.

\paragraph{Quality without utility in standard XAI.}
\label{sec:related-quality-utility}
A growing literature finds that visual XAI outputs frequently do not improve decision-making and sometimes harm it. \citet{Bansal2021DoesTW} report that explanations increase acceptance of incorrect AI recommendations without lifting team performance; \citet{Jesus2021HowCI} find LIME/SHAP lead to lower accuracy than raw data alone; \citet{im2023evaluating} extend this to oracle saliency; \citet{schemmer2022should} show that explanations do not produce appropriate reliance. The mechanism is overreliance \citep{bucinca2021trust, Bucinca2020ProxyTasks, ChenV2023understanding}. In the placebic-explanation thread, \citet{shymanski2025beyond, Shymanski2025NotAE} show that users rate placebic and actionable explanations as equally satisfying; \citet{Ajwani2024LLMGeneratedBE} and \citet{fan2026persuades} document how LLM explanations sustain trust in incorrect outputs through fluency and framing; \citet{Spillner2026NotAllTrust} show that self-reported trust dissociates from behavioural reliance. The constructs \emph{quality} \citep{Nauta2023SystematicReview, Naveed2024OverviewXAI} and \emph{usefulness} \citep{doshi2017towards, jacovi2021formalizing} are explicitly distinguished but rarely jointly measured.

\paragraph{LLM-generated NLEs from XAI outputs.}
\label{sec:related-xai-nle}
Recent systems convert SHAP attributions into prose: XAIstories \citep{martens2024xaistories}, Explingo \citep{zytek2024explingo}, ContextualSHAP \citep{Dwiyanti2025ContextualSHAP}, GraphXAIN for graph models \citep{cedro2025graphxain}, the social-science-grounded iLLuMinaTE \citep{swamy2025illuminate}, and the time-series--specific XForecast \citep{aksu2024xforecast}. All evaluate exclusively on quality proxies -- subjective ratings, comprehension surveys, automated grading. \citet{fabian2025fromxaitostories} follow the same convention, reaching 4--5/5 quality across a factorial design.Whether high NLE quality translates to downstream usefulness has not been tested.

\paragraph{LLM-as-judge.}
\label{sec:related-llm-judge}
LLM judges are now standard for open-ended generation evaluation where BLEU/ROUGE fail \citep{liu2023geval, zheng2023judging, Gu2025LLMJudge}, despite documented systematic biases -- position bias, verbosity bias, and self-preference (a judge favouring outputs from its own model family) \citep{zheng2023judging, gu2024llmasjudge}. \citet{liu2023geval} introduced G-Eval, whose chain-of-thought protocol we adapt below. In the XAI domain, \citet{BombasseiDeBona2024Evaluating} show that LLM judges replicate human conclusions on coarse-grained qualities but are weaker on numerical verification.

\paragraph{Positioning.}
\label{sec:related-positioning}
We ask whether the quality-versus-usefulness gap documented for visual XAI also holds for LLM-generated NLEs. Holding NLE quality constant at the levels established in \citet{fabian2025fromxaitostories}, we test five distinct usefulness constructs under controlled conditions.

\section{Methodology}
\label{sec:method}

We build on the prediction-to-NLE pipeline of \citet{fabian2025fromxaitostories}, fixed here to the configuration that study identified as highest-performing and most efficient: XGBoost + SHAP TreeExplainer + zero-shot prompting, with GPT-4o and DeepSeek-R1 as generators and judges. Holding quality constant at the levels established there makes NLE usefulness the sole variable of interest.

\begin{algorithm}[H]
\caption{NLE Generation and Evaluation}
\label{alg:pipeline}
\small
\begin{algorithmic}[1]
\Require Test set $\mathcal{D}_\text{test}$ ($N{=}60$), XGBoost $f$, SHAP explainer $\chi$
\Require Generators $\mathcal{G}{=}\{\text{GPT-4o, DeepSeek-R1}\}$
\Require Judges $\mathcal{J}{=}\{\text{GPT-4o, DeepSeek-R1}\}$
\Ensure Judgment corpus $\mathcal{R}$
\Statex \textit{// Phase 1: NLE Generation (shared across experiments)}
\For{all instances $\mathbf{x}_t \in \mathcal{D}_\text{test}$}
    \State $\hat{y}_t \gets f(\mathbf{x}_t)$;\;\; $\boldsymbol{\phi}_t \gets \chi(f, \mathbf{x}_t)$
    \For{all generators $g \in \mathcal{G}$}
        \State $E_t^g \gets \textsc{ZeroShot}(g;\, \mathbf{x}_t, \hat{y}_t, \boldsymbol{\phi}_t)$
    \EndFor
\EndFor
\Statex \textit{// Phase 2: Evaluation (per experiment)}
\For{experiments $e$, conditions $c \in \mathcal{C}_e$, instances $\mathbf{x}_t$, judges $j \in \mathcal{J}$}
    \State Store $\textsc{Judge}(j;\, \textsc{BuildPrompt}(\mathbf{x}_t, c, E_t))$ in $\mathcal{R}$
\EndFor
\State \Return $\mathcal{R}$
\end{algorithmic}
\end{algorithm}

\begin{figure*}[!t]
\centering
\resizebox{0.7\textwidth}{!}{

\definecolor{ExpE1}{HTML}{2E8B9A}
\definecolor{ExpE2}{HTML}{D98A3D}
\definecolor{ExpE3}{HTML}{7E5FA8}
\definecolor{ExpE4}{HTML}{B94A55}
\definecolor{ExpE5}{HTML}{C85A8A}

\definecolor{RowFeatures}{HTML}{708090}
\definecolor{RowPred}{HTML}{7E5FA8}
\definecolor{RowSHAP}{HTML}{3D7A8A}
\definecolor{RowMetrics}{HTML}{B85A7A}
\definecolor{RowNLE}{HTML}{D98A3D}

\definecolor{RowBg}{HTML}{F8F4EC}
\definecolor{InputOuterBg}{HTML}{EDF1F5}
\definecolor{BadgeBg}{HTML}{EDE4F5}
\definecolor{BadgeFrame}{HTML}{8878AA}

\begin{tikzpicture}[
    font=\sffamily\footnotesize,
    >={Latex[length=2.5mm,width=2mm]},
    inputbox/.style={
        rectangle, rounded corners=6pt,
        draw=ThesisLine, fill=InputOuterBg,
        line width=1pt
    },
    pill/.style={
        rectangle, rounded corners=3pt,
        fill=#1, draw=#1, text=white,
        line width=0.6pt,
        minimum width=2.1cm, minimum height=0.7cm,
        align=center,
        font=\sffamily\footnotesize\bfseries
    },
    rowbox/.style={
        rectangle, rounded corners=3pt,
        fill=RowBg, draw=ThesisLine,
        line width=0.5pt,
        minimum width=5cm, minimum height=0.85cm,
        text width=4.7cm,
        inner sep=4pt,
        align=left,
        font=\sffamily\scriptsize\ttfamily,
        text=ThesisNavy
    },
    badge/.style={
        rectangle, rounded corners=2pt,
        fill=BadgeBg, draw=BadgeFrame, text=BadgeFrame,
        font=\sffamily\tiny\bfseries,
        inner xsep=4pt, inner ysep=2pt,
        line width=0.4pt,
        align=center,
        text width=1.2cm
    },
    judge/.style={
        rectangle, rounded corners=6pt,
        fill=ThesisNavy, text=white,
        draw=ThesisNavy, line width=1pt,
        minimum width=2.4cm, minimum height=2.6cm,
        align=center,
        font=\sffamily\small\bfseries
    },
    expbody/.style={
        rectangle, rounded corners=0pt,
        fill=white, draw=#1,
        line width=0.9pt,
        text width=4.6cm,
        inner sep=5pt,
        align=left,
        font=\sffamily\scriptsize,
        text=ThesisNavy
    },
    expheader/.style={
        rectangle, rounded corners=0pt,
        fill=#1, draw=#1, text=white,
        line width=0.9pt,
        text width=4.6cm,
        inner xsep=7pt, inner ysep=3pt,
        align=left,
        font=\sffamily\footnotesize\bfseries
    },
    bucket/.style={
        rectangle, rounded corners=2pt,
        fill=#1!15, draw=#1!70, text=#1,
        line width=0.4pt,
        inner xsep=3pt, inner ysep=1.5pt,
        font=\sffamily\scriptsize
    },
    arr/.style={-{Latex[length=3mm,width=2.5mm]}, line width=1pt, draw=ThesisNavy!75}
]

\newcommand{\chip}[2]{\tikz[baseline=(X.base)]{\node[bucket=#1, inner sep=2pt] (X) {#2};}}

\def\xPill{0.2}
\def\xRow{2.6}
\def\xBadge{7.9}
\def\xJudge{11.5}
\def\xExp{13.3}

\def\rowgap{0.4}


\def\cardgap{3mm}

\node[expheader=ExpE1, anchor=north west] (e1-h) at (\xExp, 0) {E1 Closeness};
\node[expbody=ExpE1, anchor=north west] (e1-b) at (e1-h.south west)
    {Error bucket?\\[3pt]
     \chip{ExpE1}{small}\ \chip{ExpE1}{med}\ \chip{ExpE1}{large}\ \chip{ExpE1}{very large}};

\node[expheader=ExpE2, anchor=north west] (e2-h) at ([yshift=-\cardgap]e1-b.south west) {E2 Placebic};
\node[expbody=ExpE2, anchor=north west] (e2-b) at (e2-h.south west)
    {Error bucket?\\[3pt]
     \chip{ExpE2}{small}\ \chip{ExpE2}{med}\ \chip{ExpE2}{large}\ \chip{ExpE2}{very large}};

\node[expheader=ExpE3, anchor=north west] (e3-h) at ([yshift=-\cardgap]e2-b.south west) {E3 Counterfactual};
\node[expbody=ExpE3, anchor=north west] (e3-b) at (e3-h.south west)
    {Direction after perturbation?\\[3pt]
     \chip{ExpE3}{UP}\ \chip{ExpE3}{DOWN}\ \chip{ExpE3}{SIMILAR}};

\node[expheader=ExpE4, anchor=north west] (e4-h) at ([yshift=-\cardgap]e3-b.south west) {E4 Mental Model};
\node[expbody=ExpE4, anchor=north west] (e4-b) at (e4-h.south west)
    {Error bucket?\ {\scriptsize\itshape (after 5 training examples)}\\[3pt]
     \chip{ExpE4}{small}\ \chip{ExpE4}{med}\ \chip{ExpE4}{large}\ \chip{ExpE4}{very large}};

\node[expheader=ExpE5, anchor=north west] (e5-h) at ([yshift=-\cardgap]e4-b.south west) {E5 Selective Reliance};
\node[expbody=ExpE5, anchor=north west] (e5-b) at (e5-h.south west)
    {Reliable or unreliable?\\[3pt]
     \chip{ExpE5}{reliable}\ \chip{ExpE5}{unreliable}};

\node[rectangle, rounded corners=4pt,
      fill=PromptOutputBg, draw=PromptOutputFrame, line width=0.7pt,
      text=PromptOutputTitle, font=\sffamily\scriptsize,
      text width=4.6cm, align=center,
      inner sep=4pt,
      anchor=north west] (conf) at ([yshift=-5mm]e5-b.south west)
    {All experiments also ask:\\ \textbf{How confident are you? (1--5)}};

\coordinate (r-mid) at ($(e1-h.north)!0.5!(conf.south)$);


\def\LeftHalf{3.15}

\node[rowbox, anchor=north west] (r-feat)
    at ($(\xRow, 0 |- r-mid) + (0, \LeftHalf)$)
    {lag\_1: 166.27,\ \ lag\_2: 162.28,\\
     lag\_3: 189.77,\ \ \dots,\\
     lag\_7: 129.44,\ \ weekofyear: 42,\\
     holidays: 0};
\node[pill=RowFeatures, anchor=west]
    (p-feat) at (\xPill, 0 |- r-feat.center) {Features};
\node[badge, anchor=west]
    (b-feat) at (\xBadge, 0 |- r-feat.center) {always\\present};

\node[rowbox, anchor=north west, minimum height=0.7cm,
      font=\sffamily\small\ttfamily\bfseries]
    (r-pred) at ([yshift=-\rowgap cm]r-feat.south west) {168.76 kWh};
\node[pill=RowPred, anchor=west]
    (p-pred) at (\xPill, 0 |- r-pred.center) {Prediction};
\node[badge, anchor=west]
    (b-pred) at (\xBadge, 0 |- r-pred.center) {always\\present};

\node[rowbox, anchor=north west]
    (r-shap) at ([yshift=-\rowgap cm]r-pred.south west)
    {lag\_6: $-14.26$,\ \ lag\_5: $-9.74$,\\
     lag\_7: $+6.14$,\ \ \dots};
\node[pill=RowSHAP, anchor=west]
    (p-shap) at (\xPill, 0 |- r-shap.center) {X (SHAP)};

\node[rowbox, anchor=north west, minimum height=0.7cm]
    (r-met) at ([yshift=-\rowgap cm]r-shap.south west)
    {R\textsuperscript{2} = 0.686,\ \ MAE = 20.55,\ \ RMSE = 25.04};
\node[pill=RowMetrics, anchor=west]
    (p-met) at (\xPill, 0 |- r-met.center) {T (Metrics)};

\node[rowbox, anchor=north west, text width=4.7cm, minimum height=1.7cm,
      font=\sffamily\scriptsize]
    (r-nle) at ([yshift=-\rowgap cm]r-met.south west)
    {\textbf{Key Influences:} The prediction
     (\textbf{168.76 kWh}) is below the average
     baseline (177.36 kWh) due to lower past
     consumption. The largest downward impact
     came from energy use 6 weeks prior\dots};
\node[pill=RowNLE, anchor=west]
    (p-nle) at (\xPill, 0 |- r-nle.center) {E (NLE)};

\begin{scope}[on background layer]
    \node[inputbox, fit=(p-feat)(r-feat)(p-nle)(r-nle)(b-feat)(b-pred),
          inner xsep=8pt, inner ysep=10pt] (inputbox-bg) {};
\end{scope}

\node[font=\sffamily\small\bfseries, text=ThesisNavy, anchor=south west]
    (input-title) at ([xshift=3pt,yshift=2pt]inputbox-bg.north west) {LLM Judge Input};


\node[judge, anchor=center]
    (judge) at (\xJudge, 0 |- r-mid)
    {LLM Judge\\[3pt]{\footnotesize\mdseries GPT-4o}\\{\footnotesize\mdseries DeepSeek-R1}};

\draw[arr] (inputbox-bg.east |- r-mid) -- (judge.west);
\draw[arr] (judge.east) -- ($(e3-h.west |- r-mid) + (-0.12cm, 0)$);

\end{tikzpicture}}
\caption{Experimental framework. Left: the LLM judge receives up to five information pieces per instance -- features and prediction are always present; SHAP values (X), model metrics (T), and the NLE (E) vary by condition. Right: the five downstream tasks with their response categories and confidence rating (1--5).}
\label{fig:experiment_overview}
\end{figure*}
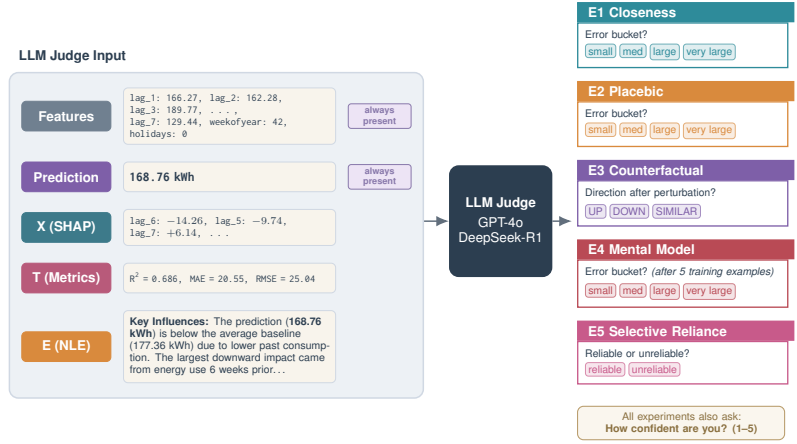

Algorithm~\ref{alg:pipeline} gives an overview of the two-phase design shared by all five experiments. Phase~1 generates 120 NLEs once (60 instances $\times$ 2 generators), reused across experiments. Phase~2 then runs five downstream-task experiments, each ablating a subset of $\{$features, SHAP, metrics, NLE$\}$ -- the condition set $\mathcal{C}_e$ -- to isolate the marginal contribution of the NLE; Figure~\ref{fig:experiment_overview} visualises this framework, and the individual tasks and conditions $\mathcal{C}_e$ are introduced in \S\ref{sec:exp1}--\S\ref{sec:exp5}. The corpus $\mathcal{R}$ contains 2{,}730 judgments analysed via the mixed-effects models of \S\ref{sec:method-stats}.

\begin{figure}[H]
\centering
\includegraphics[width=\columnwidth]{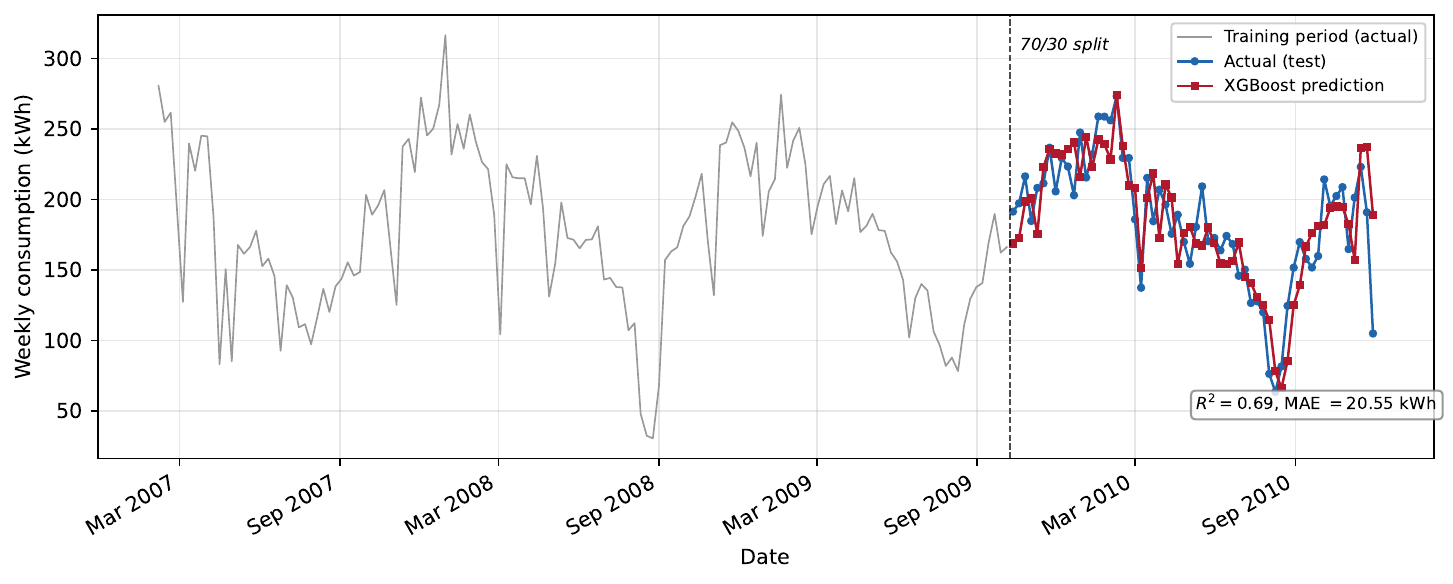}
\caption{XGBoost one-step-ahead predictions vs.\ actual weekly consumption (kWh). Grey: training period; coloured: test set.}
\label{fig:pred_vs_actual}
\end{figure}

\subsection{Dataset and Prediction Model}
\label{sec:method-domain}

We use the UCI Individual Household Electric Power Consumption dataset \citep{hebrail2012individual} -- 2{,}075{,}259 minute-level measurements from a household near Paris (Dec 2006--Nov 2010), resampled to weekly granularity. Following \citet{fabian2025fromxaitostories} we engineer nine features: seven autoregressive lags (\texttt{lag\_1}--\texttt{lag\_7}, prior weeks' consumption in kWh), ISO week number, and French public-holiday count per week. After dropping rows with NaN lag values the dataset contains 200 weekly observations; a chronological 70/30 split yields 140 training and 60 test instances (full preprocessing in Appendix~\ref{app:experimental}). The domain was chosen for interpretability: consumption is governed by mechanisms (winter heating, holidays, weekly routines) that non-technical readers can reason about from world knowledge alone -- a transparent setting in which, if NLEs cannot help, the prospects elsewhere are limited.

XGBoost \citep{chen2016xgboost} (the model $f$ of Algorithm~\ref{alg:pipeline}) achieves $R^2 = 0.69$, MAE $= 20.55$~kWh, RMSE $= 25.04$~kWh on the test set (Figure~\ref{fig:pred_vs_actual}; absolute percentage errors span 0.3--80.2\%). All predictions are one-step-ahead (true preceding lags as input, not recursive outputs). For each test instance $\mathbf{x}_t$ we obtain a prediction $\hat{y}_t = f(\mathbf{x}_t)$ and SHAP TreeExplainer attributions \citep{lundberg2017unified, lundberg2020local} $\boldsymbol{\phi}_t = \chi(f, \mathbf{x}_t)$, decomposing $\hat{y}_t$ into additive per-feature contributions relative to a base value. The full hyperparameter configuration, training procedure, and SHAP computation code are provided in Appendix~\ref{app:experimental}.

\subsection{NLE Generation and Evaluation}
\label{sec:method-nle}

\paragraph{Fixed pipeline.} Each factor is justified empirically by \citet{fabian2025fromxaitostories}: XGBoost yielded the highest NLE quality ($d = 0.48$--$0.82$ over alternatives); SHAP and LIME produced equivalent quality ($\omega^2 = 0.02$), and TreeExplainer additionally offers exact Shapley values; zero-shot achieved near-equivalent quality to self-consistency (4.36 vs.\ 4.50/5) at $7\times$ lower token cost; GPT-4o (4.29/5) and DeepSeek-R1 (4.66/5) were the two highest-quality generators (Llama-3-8B was excluded for low quality and inconsistent formatting). GPT-4o is accessed via Azure OpenAI (deployment \texttt{gpt-4o}, API version \texttt{2024-10-21}); DeepSeek-R1 \citep{guo2025deepseekr1} via OpenRouter (\texttt{deepseek/deepseek-r1}). The two span dense vs.\ Mixture-of-Experts architectures and RLHF vs.\ reasoning-focused RL training paradigms.

\paragraph{Generation.} For each test instance $\mathbf{x}_t$ and each generator $g \in \mathcal{G}$, we obtain one NLE $E_t^g$ at sampling temperature $\tau = 1.0$, yielding 120 NLEs ($60 \times 2$). The generation prompt is taken verbatim from \citet{fabian2025fromxaitostories}: a system prompt plus a human message supplying domain context, performance metrics, the prediction, raw features, and SHAP values sorted by absolute magnitude. Output is capped at six bullets and 200 words (controlling verbosity bias; \citealp{gu2024llmasjudge}); full prompt, worked instance, and example output in Appendix~\ref{sec:appendix-nle-prompt}. The same 120 NLEs are reused across E1--E4; E5 uses separately generated NLEs over poisoned inputs.

\paragraph{Judge design.} Each of the five experiments is a downstream decision-making study in which an LLM judge -- standing in for the non-expert reader the NLE pipeline ultimately targets -- is shown the XGBoost prediction together with a varying subset of $\{$raw features, SHAP attributions, performance metrics, NLE$\}$ (the condition $\mathcal{C}_e$) and asked to perform an experiment-specific task (e.g., classify the prediction's percentage-error bucket); it also reports a 1--5 Likert confidence. The two LLMs that produced the NLEs (GPT-4o, DeepSeek-R1) also act as judges, $\mathcal{J} = \mathcal{G}$ in Algorithm~\ref{alg:pipeline}: NLE conditions yield four generator--judge combinations per instance, no-NLE conditions yield two (judges only, since there is no NLE and hence no generator dimension); cross-family combinations cancel self-preference bias \citep{zheng2023judging, gu2024llmasjudge}, and judge identity is carried as a fixed effect in every statistical model. Judge prompts are G-Eval-style with chain-of-thought evaluation steps, run at sampling temperature $\tau = 0.0$ for deterministic evaluation; the structured-JSON response carries the answer, the confidence rating, and a one-sentence reasoning. The exact prompt for each experiment is reproduced verbatim in Appendix~\ref{sec:appendix-prompts}, alongside a worked example of a complete judge human message for one instance. Whether the NLE helps is then answered statistically: for each experiment we compare task accuracy across information conditions via the mixed-effects models of \S\ref{sec:method-stats}. Figure~\ref{fig:experiment_overview} summarises the framework -- the left panel maps each condition to the information pieces in the judge's human message (features + prediction always present; SHAP, metrics, and NLE toggled), the right panel lists the five tasks.

\subsection{Statistical Analysis}
\label{sec:method-stats}

Within each experiment, the same instances appear under every condition (within-instance repeated measures), so judgments on the same instance are not independent. We therefore use mixed-effects models throughout, with a random intercept per instance to absorb the residual variance shared by judgments on the same item.

\paragraph{Accuracy.} Binary correctness $y_{ij} \in \{0,1\}$ for judgment $i$ on instance $j$ is modelled with a generalised linear mixed model (GLMM) using a binomial family and logit link, fitted by maximum likelihood via \texttt{lme4}'s \texttt{glmer} \citep{bates2015lme4}:
\begin{equation}
\label{eq:glmm}
\begin{aligned}
\text{logit}\big(P(y_{ij}{=}1)\big) ={} & \beta_0 + \boldsymbol{\beta}_{c}^{\!\top} C_{ij} + \beta_{j} J_{ij} \\
& {}+ \boldsymbol{\beta}_{c\times j}^{\!\top}(C{\times}J)_{ij} + u_j,
\end{aligned}
\end{equation}
where $C_{ij}$ is a dummy-coded vector of condition indicators, $J_{ij}$ the judge indicator (GPT-4o vs.\ DeepSeek-R1), $(C{\times}J)_{ij}$ their interaction (carried whenever the model converges), and $u_j \sim \mathcal{N}(0, \sigma_u^2)$ the random intercept for instance $j$. The omnibus condition effect is tested via the likelihood-ratio test (LRT) between \eqref{eq:glmm} and a reduced model that drops both $C$ and $C{\times}J$. Pairwise contrasts are extracted as Wald $z$-tests on estimated marginal means (EMMs) with Holm-Bonferroni correction \citep{lenth2024emmeans, holm1979simple}. The primary effect size is the odds ratio $\text{OR} = \exp(\beta_c)$: $\text{OR} = 1$ indicates no effect, $\text{OR} > 1$ multiplies the odds of a correct judgment by that factor, and the 95\% confidence interval (CI) indicates precision.

\paragraph{Confidence.} Ordinal confidence (1--5 Likert) is modelled with a cumulative link mixed model (CLMM) via the \texttt{ordinal} package \citep{christensen2019ordinal}:
\begin{equation}
\label{eq:clmm}
\begin{aligned}
P(\text{conf}_{ij} \leq k) = \text{logit}^{-1}\big( & \theta_k - \boldsymbol{\beta}_{c}^{\!\top} C_{ij} \\
& {}- \beta_{j} J_{ij} - u_j\big),
\end{aligned}
\end{equation}
$k = 1,\ldots,4$, with $\theta_1 < \theta_2 < \theta_3 < \theta_4$ the thresholds separating the five Likert categories. Under the proportional-odds assumption, a positive $\beta_c$ multiplies the odds of reporting $\geq k$ rather than $<k$ by $\exp(\beta_c)$ at every threshold. To diagnose miscalibrated trust we additionally fit \eqref{eq:clmm} with a correctness$\times$condition interaction (the \emph{overconfidence} model): a positive interaction term indicates that confidence rises with the condition even among incorrect judgments. When the interaction model fails to converge, the additive model is reported as the primary fit and the interaction is evaluated separately via LRT.

\paragraph{Bayesian analysis and ROPE.} A frequentist non-rejection does not by itself establish the \emph{absence} of an effect. For every null accuracy result we therefore fit a Bayesian analogue of \eqref{eq:glmm} via \texttt{brms} \citep{burkner2017brms} with weakly informative priors -- $\mathcal{N}(0, 1.5)$ on fixed effects \citep{gelman2008weakly} and half-$t_3(0, 2.5)$ on the random-intercept standard deviation -- using 4 chains of 4{,}000 iterations each. We report posterior odds ratios with 95\% credible intervals (CrI) and the proportion of posterior mass inside a Region of Practical Equivalence (ROPE) -- a range of effect sizes we treat as ``effectively no effect''. We fix the ROPE on the log-odds scale as $[-0.18, +0.18]$ ($\text{OR} \in [0.84, 1.20]$, roughly $\pm 5$~pp at a 40\% baseline). A posterior with $\geq 95\%$ of its mass inside the ROPE supports practical equivalence to zero; lower proportions quantify how much posterior mass falls outside the region of triviality.

\paragraph{Sensitivity checks.} For every primary model we run judge-specific GLMMs, generator and same-family-bias tests (NLE conditions only), random-slope specifications, and Friedman / Wilcoxon nonparametric backups. None of these alters the primary conclusions; full results are in Appendix~\ref{sec:appendix-stats}.

\paragraph{Reporting plan.} Each experiment (\S\ref{sec:exp1}--\S\ref{sec:exp5}) follows the same template: \emph{Design} states the task and conditions; \emph{Results} reports descriptive accuracy and confidence, the omnibus LRT, the NLE OR with 95\% CI and Holm-corrected $p$, the confidence CLMM, and (for nulls) the Bayesian posterior OR. Full coefficient tables, pairwise contrasts, calibration outputs, diagnostics, sensitivity analyses, and judge prompts are in Appendices~\ref{sec:appendix-stats} and~\ref{sec:appendix-prompts}; \S\ref{sec:discussion} synthesises the five experiments.

\subsection{Experiment Selection}
\label{sec:method-selection}

Explanation \emph{usefulness} is not a single construct: the XAI evaluation literature distinguishes several facets, each requiring its own operationalisation \citep{doshi2017towards, hase2020evaluating, Bansal2021DoesTW, Nauta2023SystematicReview}. We chose one experiment per facet, plus one mechanism-diagnostic control. E1 (\emph{forward simulatability}) and E3 (\emph{counterfactual simulatability}) together cover the two simulatability operationalisations of \citet{doshi2017towards} and \citet{hase2020evaluating}: can the judge predict the model's past performance, and its response to feature perturbations? E4 (\emph{mental model transfer}) tests the learning facet \citep{Bansal2021DoesTW, Bucinca2020ProxyTasks}: does prior NLE exposure produce understanding that transfers to a test instance without NLE? E5 (\emph{selective reliance under distribution shift}) tests the deployment-critical question of whether the judge appropriately distrusts predictions on out-of-distribution inputs \citep{parasuraman1997humans, lee2004trust, bucinca2021trust}. E2 (\emph{placebic content/presence control}) is not itself a usefulness measure but a mechanism diagnostic \citep{langer1978mindlessness, shymanski2025beyond}: it separates whether any NLE-induced effect comes from the NLE's content or merely from the presence of additional text. Together, the four constructs (assessment of past performance, prediction of future behaviour, learning, selective reliance) span the dominant downstream-usefulness criteria in the XAI literature, and the placebic control isolates the mechanism behind whatever pattern they yield.

\section{E1: Forward Simulatability}
\label{sec:exp1}

\subsection{Design}
\label{sec:exp1-design}

\paragraph{Task.} The judge classifies the XGBoost prediction's absolute percentage error into one of four ordered buckets -- \textit{small} ($<5\%$), \textit{medium} ($5$--$15\%$), \textit{large} ($15$--$30\%$), \textit{very large} ($\geq 30\%$) -- and reports a 1--5 Likert confidence.

\begin{table}[H]
\centering
\small
\resizebox{\columnwidth}{!}{%
\begin{tabular}{lcccc}
\toprule
\textbf{Condition} & \textbf{Features} & \textbf{SHAP (X)} & \textbf{Metrics (T)} & \textbf{NLE (E)} \\
\midrule
Baseline & \checkmark & & & \\
X        & \checkmark & \checkmark & & \\
T        & \checkmark & & \checkmark & \\
X+T      & \checkmark & \checkmark & \checkmark & \\
E+X+T    & \checkmark & \checkmark & \checkmark & \checkmark \\
\bottomrule
\end{tabular}%
}
\caption{E1 information conditions. All include prediction, features, model name, and domain context. The key contrast is X+T vs.\ E+X+T.}
\label{tab:e1-conditions}
\end{table}

\paragraph{Information conditions.} Each instance is evaluated under five conditions (Table~\ref{tab:e1-conditions}). All include the prediction, raw features, model name, and domain context; the structured bundles -- SHAP values (\textbf{X}) and performance metrics (\textbf{T}: MAE, RMSE, $R^2$) -- are added incrementally.

Standalone NLE conditions (E, E+X, E+T) are deliberately excluded: NLEs were generated with full context and narrate SHAP and metrics in prose, so including one without the corresponding structured data would leak that information through the narrative. The only clean NLE test is therefore E+X+T vs.\ X+T. The four no-NLE conditions yield $4\times 60\times 2 = 480$ judgments; E+X+T uses the $2\times 2$ generator--judge design for $60\times 4 = 240$. Total $N = 720$. Ground-truth bucket distribution across instances: 12 / 29 / 16 / 3 (small/medium/large/very-large).

\subsection{Results}
\label{sec:exp1-results}

\begin{table}[H]
\centering
\small
\begin{tabular}{lcS[table-format=1.3]c}
\toprule
\textbf{Condition} & \textbf{$n$} & {\textbf{Accuracy}} & \textbf{Mean Conf.} \\
\midrule
Baseline & 120 & 0.358 & 3.33 \\
X        & 120 & 0.383 & 3.41 \\
T        & 120 & 0.392 & 3.50 \\
X+T      & 120 & 0.417 & 3.81 \\
E+X+T    & 240 & 0.392 & 3.92 \\
\bottomrule
\end{tabular}
\caption{E1 accuracy and mean confidence by condition.}
\label{tab:e1-results}
\end{table}

\paragraph{Accuracy.} The GLMM shows no omnibus condition effect (LRT $\chi^2(4) = 1.58$, $p = .81$); the key NLE contrast E+X+T vs.\ X+T yields $\text{OR} = 0.83$ $[0.47, 1.48]$, $p = .53$ -- the NLE adds nothing beyond the structured data it narrates. All pairwise contrasts are non-significant after Holm correction. Bayesian posterior for E+X+T vs.\ Baseline: $\text{OR} = 1.25$ $[0.69, 2.18]$, ROPE 34.8\% -- consistent with a small, non-significant effect.

\paragraph{Confidence.} A CLMM yields a highly significant condition effect (LRT $\chi^2(4) = 179.1$, $p < 2.2 \times 10^{-16}$). Holm-corrected pairwise contrasts show every information-bearing condition (X+T, E+X+T) differs from every non-information one (Baseline, X, T) at $p < .0001$; the NLE-specific increment over X+T is not significant ($p = .62$). The correctness$\times$condition interaction is significant: among \emph{wrong} predictions, mean confidence rises from 3.46 (Baseline) to 3.90 (E+X+T). Full wrong/correct breakdown in Appendix~\ref{sec:appendix-stats}.

\section{E2: Placebic Control}
\label{sec:exp2}

\subsection{Design}
\label{sec:exp2-design}

The task is identical to E1 (classify error magnitude). All three conditions include features, SHAP, and metrics, and differ only in NLE source: \textit{Baseline} (no NLE), \textit{Real NLE} (NLE generated for this instance), and \textit{Placebo NLE} (NLE generated for a \emph{different} instance, drawn from a random derangement -- a permutation with no fixed points -- so each placebic NLE describes a different prediction context).

\subsection{Results}
\label{sec:exp2-results}

\begin{table}[H]
\centering
\small
\begin{tabular}{lcS[table-format=1.3]c}
\toprule
\textbf{Condition} & \textbf{$n$} & {\textbf{Accuracy}} & \textbf{Mean Conf.} \\
\midrule
Baseline     & 120 & 0.450 & 3.81 \\
Real NLE     & 240 & 0.479 & 3.95 \\
Placebo NLE  & 240 & 0.458 & 3.90 \\
\bottomrule
\end{tabular}
\caption{E2 accuracy and confidence by condition.}
\label{tab:e2-results}
\end{table}

\paragraph{Accuracy.} The omnibus condition effect is null ($\chi^2(2) = 0.68$, $p = .71$); Holm-corrected pairwise contrasts all yield $p = 1.0$ -- Real vs.\ Baseline $\text{OR} = 1.26$, Placebo vs.\ Baseline $\text{OR} = 1.07$, and the critical Placebo vs.\ Real $\text{OR} = 0.85$. Bayesian posteriors corroborate: Real $\text{OR} = 1.24$ $[0.68, 2.28]$, Placebo $\text{OR} = 1.06$ $[0.58, 1.89]$. The three conditions are indistinguishable on accuracy.

\paragraph{Confidence.} The additive CLMM (the interaction model produced near-zero random-intercept variance) yields Real vs.\ Baseline $1.10$, $p = .002$; Placebo vs.\ Baseline $0.68$, $p = .051$; Real vs.\ Placebo $\Delta = -0.42$, $p = .12$. Both NLE types elevate confidence relative to Baseline; the two do not differ.

\section{E3: Counterfactual Simulatability}
\label{sec:exp3}

\subsection{Design}
\label{sec:exp3-design}

For each test instance we perturb the three most influential features (by absolute SHAP value): lag features by $\pm 25\%$ of their value (random sign), \texttt{weekofyear} by $\pm 13$ weeks (clamped to $[1,52]$), \texttt{holiday\_week\_count} by $+2$. The XGBoost model is re-evaluated, and the judge predicts whether the output is \textit{higher}, \textit{lower}, or \textit{similar} ($<5\%$ change). Both conditions include features, SHAP, and metrics; they differ only in whether the NLE for the original instance is shown. Total $N = 360$; ground truth: 28 higher, 24 lower, 8 similar.

\subsection{Results}
\label{sec:exp3-results}

\begin{table}[H]
\centering
\small
\begin{tabular}{lcS[table-format=1.3]c}
\toprule
\textbf{Condition} & \textbf{$n$} & {\textbf{Accuracy}} & \textbf{Mean Conf.} \\
\midrule
SHAP only (X)     & 120 & 0.533 & 3.97 \\
NLE + SHAP (E+X)  & 240 & 0.579 & 4.00 \\
\bottomrule
\end{tabular}
\caption{E3 direction accuracy by condition.}
\label{tab:e3-results}
\end{table}

\paragraph{Accuracy.} Both conditions clear the 33.3\% chance rate. The X vs.\ E+X contrast yields $\text{OR} = 1.37$ $[0.79, 2.37]$, $p = .27$ (Bayesian posterior $\text{OR} = 1.36$ $[0.76, 2.46]$). Per-class recall: \textit{similar} is hardest (19--25\%), \textit{higher} and \textit{lower} comparable (55--64\%).

\paragraph{Confidence.} Confidence is at ceiling ($\sim$4.0) in both conditions (CLMM $z = 0.70$, $p = .49$), leaving no room for NLE-induced inflation. A significant correctness$\times$condition interaction on calibration ($p = .019$) suggests NLEs marginally improve confidence--accuracy tracking without affecting accuracy, but overconfidence among incorrect predictions remains extreme ($\sim$71--77\%).

\section{E4: Mental Model Transfer}
\label{sec:exp4}

\subsection{Design}
\label{sec:exp4-design}

A sliding-window in-context paradigm: for each test position $i \geq 5$ (55 positions), the judge first sees five preceding instances ($i-5$ to $i-1$) as training examples with features, SHAP, metrics, prediction, true value, and correct error bucket; then classifies the bucket of instance $i$ -- with features, SHAP, prediction, but no true value -- and \emph{no NLE}. The two conditions differ only in whether the five training examples include NLEs; the test instance never does.

\subsection{Results}
\label{sec:exp4-results}

\begin{table}[H]
\centering
\small
\begin{tabular}{lcS[table-format=1.3]c}
\toprule
\textbf{Condition} & \textbf{$n$} & {\textbf{Accuracy}} & \textbf{Mean Conf.} \\
\midrule
Baseline       & 110 & 0.364 & 3.57 \\
NLE training   & 220 & 0.405 & 3.57 \\
\bottomrule
\end{tabular}
\caption{E4 accuracy by condition.}
\label{tab:e4-results}
\end{table}

\paragraph{Accuracy.} The GLMM yields a directional but non-significant gain ($\text{OR} = 1.86$ $[0.75, 4.61]$, $p = .18$; Bayesian posterior $\text{OR} = 1.86$ $[0.76, 4.73]$). The wide interval reflects extreme between-instance difficulty (random-intercept SD $= 5.20$); trial order is null ($p = .76$), ruling out learning-over-time. A design caveat applies: the sliding window induces 80\% overlap between adjacent training sets, producing temporal autocorrelation only imperfectly absorbed by the instance random intercept.

\paragraph{Confidence.} Confidence is identical across conditions (mean 3.57, $p = .97$): when NLEs appear only in training, no confidence inflation occurs at test time.

\section{E5: Selective Reliance}
\label{sec:exp5}

\subsection{Design}
\label{sec:exp5-design}
\paragraph{Data manipulation.} The 60 test instances form the \textit{baseline} condition. \textit{Out-of-distribution} (OOD) variants are constructed by season-dependent lag poisoning -- cold-season weeks are pushed low, warm-season weeks high (full code in Appendix~\ref{app:experimental}). XGBoost and TreeExplainer are then re-run on the poisoned inputs: predictions and SHAP values are internally consistent with the poisoned features but systematically misrepresent real consumption. NLEs for poisoned instances are generated separately by the same pipeline (\S\ref{sec:method-nle}) -- they faithfully describe the poisoned attributions, which is precisely the rationalisation mechanism under test.

\paragraph{Task.} The judge classifies each instance as \emph{reliable} or \emph{unreliable} (ground truth: baseline reliable, OOD unreliable). All conditions include features, SHAP, and metrics. NLE conditions use the $2\times 2$ generator--judge design ($N\times 4$); no-NLE conditions yield $N\times 2$.

\subsection{Results}
\label{sec:exp5-results}

\begin{table}[H]
\centering
\small
\begin{tabular}{llcS[table-format=1.3]c}
\toprule
\textbf{Data} & \textbf{NLE} & \textbf{$n$} & {\textbf{Accuracy}} & \textbf{Mean Conf.} \\
\midrule
Baseline & No  & 120 & 0.950 & 3.98 \\
Baseline & Yes & 240 & 0.983 & 3.99 \\
OOD      & No  & 120 & 0.300 & 3.82 \\
OOD      & Yes & 240 & 0.150 & 3.89 \\
\bottomrule
\end{tabular}
\caption{E5 accuracy and confidence by data type and NLE presence.}
\label{tab:e5-results}
\end{table}


\paragraph{Manipulation check.} Baseline instances are correctly flagged reliable at 95.0--98.3\%; OOD detection is only 15--30\% overall.

\paragraph{False reassurance.} The factorial GLMM reveals a significant poisoning $\times$ NLE interaction (LRT $\chi^2(1) = 9.84$, $p = .002$). On baseline instances NLEs marginally help (OR No NLE vs.\ NLE $= 0.31$, $p = .08$); on OOD instances NLEs significantly \emph{hurt} detection (OR $= 3.13$, $p = .0003$), halving accuracy from 30.0\% to 15.0\%. Bayesian analysis: posterior interaction $\text{OR} = 0.12$ $[0.04, 0.37]$ -- the only effect in this study whose 95\% credible interval excludes 1.

\paragraph{Confidence and calibration.} Confidence is uniformly high across all four conditions (3.82--3.99; CLMM interaction $p = .47$). Calibration is inverted in OOD (Somers' $D = -0.44$ no-NLE, $-0.37$ NLE): higher confidence is associated with \emph{lower} accuracy.

\section{Discussion}
\label{sec:discussion}

E1--E4 yield NLE-vs.-no-NLE odds ratios between $0.83$ and $1.86$, with Bayesian credible intervals all including unity: across four facets of usefulness that the XAI literature treats as distinct, we detect no accuracy benefit from the high-quality NLEs of \citet{fabian2025fromxaitostories}. The one significant accuracy effect (E5) is in the opposite direction: under distribution shift, OOD detection falls from $30\%$ to $15\%$ in the NLE's presence (interaction CrI $[0.04, 0.37]$). Confidence, in contrast, tracks NLE presence rather than accuracy: where NLEs are shown at judgment time, judges report higher confidence (E1), and E2 indicates that this elevation persists for placebic NLEs describing a different instance, with real and placebic NLEs yielding indistinguishable confidence and indistinguishable accuracy. This combination -- no detectable accuracy gain, but a content-insensitive confidence shift -- is the empirical pattern we read as the Quality-Usefulness Gap.

We suggest two tentative mechanisms, framed as hypotheses since we observe behaviour rather than cognition (our judges are LLMs; \S\ref{sec:limitations}). First, \emph{confidence inflation through peripheral processing}, plausibly accounts for E1--E2: the placebic finding -- content-free NLEs raising confidence comparably to faithful ones -- echoes the ``mindless'' processing \citet{langer1978mindlessness} identified for non-informative justifications and that \citet{shymanski2025beyond, Shymanski2025NotAE} revisited for AI explanations, and matches the Elaboration Likelihood Model's peripheral-cue prediction \citep{petty1986elaboration, kahneman2011thinking}: fluent text appears to act as a credibility cue, updating confidence without the engagement needed to check the NLE's claims against the SHAP values. Second, \emph{rationalisation of anomalous inputs}, is a candidate reading of E5, and an a~priori surprising one: frontier LLMs carry broad world knowledge about energy use and seasonality, and one might have expected an LLM narrator faced with lags far outside the seasonal norm to flag the anomaly in prose alongside reporting the SHAP attributions. In E5 this did not happen -- the pipeline narrated the misleading attributions as if sensible, and OOD detection deteriorated. Why the LLM did not exploit its world knowledge here -- whether the system prompt is too narrow, the SHAP-centric framing crowds it out, or the capability is weaker than assumed -- is an open question; an LLM narrator that reliably surfaced OOD warnings would be a natural design for a more useful pipeline. What our results show is the present-day failure mode: faithful narration of post-hoc attributions over OOD inputs appears to dampen scepticism in our setup, consistent with the adversarial-attack literature \citep{Ghorbani2019InterpretationON, slack2020fooling, Artelt2026}. The two readings also help interpret E3: SHAP attributions already encode the directional information a counterfactual prediction requires. For the judge, the prose restatement is largely a paraphrase. A common thread, conjectural rather than demonstrated, runs through all three: the surface properties optimised by the quality pipeline -- fluency, structure, accessible language -- are also those associated with peripheral credibility assessments in the standard XAI literature \citep{jacovi2021formalizing, Vasconcelos2023ExplanationsCR}; optimising for quality may, in this sense, be inadvertently optimising for persuasiveness.

Tentative as they are, these readings narrow the design space for what to do about the gap:
\begin{enumerate}[leftmargin=0pt, itemindent=0pt, listparindent=0pt, label={}]
\item \textbf{Don't ship on quality alone:} G-Eval-style scores remain a useful development-time check but do not predict deployment utility; task-based benchmarks should be reported alongside them.
\item \textbf{Evaluate at the edge:} E5 shows that an NLE system can be statistically indistinguishable from no NLE on the average input yet cause measurable harm on inputs at the edge -- adversarial detection tasks belong in pre-deployment testing, not as an afterthought.
\item \textbf{Decouple uncertainty from narrative:} If the rationalisation reading is right, a structurally separate signal of input distributional anomaly (e.g., quantile-based bounds against the training distribution) is harder to smooth over with narrative fluency, and would arguably best be displayed \emph{before} the NLE rather than inside it.
\item \textbf{Force central processing:} The peripheral-processing reading suggests the most promising interventions are not better explanations but designs that engage the reader actively: cognitive forcing functions \citep{bucinca2021trust}, devil's-advocate framings \citep{Suh2025Dont}, dissenting explanations \citep{reingold2024dissenting}, and uncertainty-aware formats \citep{Vasconcelos2023ExplanationsCR} are all candidates the standard XAI literature has developed but the LLM-NLE pipeline has yet to adopt.
\end{enumerate}

\noindent Whether any of these closes the gap in our LLM-judge setting -- let alone with human users -- is an open empirical question worth answering.

\section{Conclusion}
\label{sec:conclusion}

We tested whether the high-quality NLEs of \citet{fabian2025fromxaitostories} translate to usefulness across five tasks. They do not: no significant accuracy benefit in E1--E4 (all $p > .18$; Bayesian CrIs include $\text{OR}=1$); E2's placebic control localises the confidence gains to text presence, not content; and in E5 accuracy drops from 30\% to 15\% with the NLE ($p = .002$, CrI $[0.04, 0.37]$), consistent with false reassurance masking model failure. We call this the \textbf{Quality-Usefulness Gap}: explanation quality and usefulness can come apart, with narrative coherence plausibly acting as a trust heuristic.
\clearpage
\section*{Limitations}
\label{sec:limitations}

\paragraph{LLM judges vs.\ human users.} All usefulness experiments use LLM judges rather than human participants. This enables controlled, large-scale evaluation with a uniform judge protocol, but the cognitive mechanisms we invoke (peripheral processing, automation bias, the Elaboration Likelihood Model, System~1/System~2 distinctions) are drawn from human psychology. Our findings should be read as \emph{behavioural analogues} of the human phenomena, consistent with but not direct evidence of the underlying cognitive mechanisms. \citet{BombasseiDeBona2024Evaluating} show that LLM judges track human ratings on coarse-grained qualities but are systematically less reliable on dimensions requiring deep domain knowledge or numerical verification -- precisely the properties our usefulness tasks depend on. It is plausible that the Quality-Usefulness Gap is \emph{larger} with humans, who are typically more susceptible to fluency heuristics, but human validation studies remain the necessary next step.

\paragraph{Single domain, single pipeline.} All experiments use household electricity consumption forecasting -- a low-stakes interpretable domain -- with the SHAP TreeExplainer + XGBoost pipeline fixed by \citet{fabian2025fromxaitostories}. The mechanisms are not domain-specific, but generalisability to higher-stakes domains (healthcare, credit scoring, legal) is not established; the direction of the gap is unlikely to reverse, and consequences are likely to be larger. LIME-narrated NLE usefulness is not tested here (the factorial study found SHAP and LIME equivalent on quality), and all predictions are one-step-ahead -- recursive forecasting would degrade performance, with potentially different NLEs.

\paragraph{Compound LLM factor.} GPT-4o and DeepSeek-R1 differ on multiple axes simultaneously -- architecture (dense vs.\ MoE), parameter count, alignment strategy, access modality. A controlled future design would vary one axis at a time, but frontier LLMs are not released in such orthogonal variants. Our findings therefore speak to practical, market-available options rather than to which architectural or training property is responsible.

\paragraph{Poisoning specificity (E5).} The OOD construction pushes lag features toward seasonal extremes. Other distribution shifts -- gradual drift, covariate shift, adversarial perturbation of non-lag features -- may produce different detection patterns.

\bibliography{custom}

\appendix

\section{Experimental Details}
\label{app:experimental}

\paragraph{Preprocessing.} Missing values ($\sim$1.25\%) are filled by linear interpolation, kilowatt-minutes are converted to kilowatt-hours, and minute-level data are aggregated to weekly totals anchored on Mondays (ISO week alignment). After dropping rows with NaN lag values, the dataset contains 200 weekly observations with 9 features (\texttt{lag\_1}--\texttt{lag\_7}, \texttt{weekofyear}, \texttt{holiday\_week\_count}) and the target (weekly kWh). A chronological 70/30 split yields 140 training and 60 test instances (approximately Dec 2006 -- Aug 2009 train; Sept 2009 -- Nov 2010 test).

\begin{tcolorbox}[codebox, title={Data Loading and Feature Engineering (excerpt)}]
\begin{lstlisting}[style=pythonstyle]
def load_weekly(file_path, country="FR"):
    df = pd.read_csv(file_path, sep=";", na_values="?")
    df["DateTime"] = pd.to_datetime(df["Date"]
                                    + " " + df["Time"])
    df["Global_active_power"] = (
        df["Global_active_power"].astype(float)
          .interpolate() * (1/60)
    )
    w = df["Global_active_power"].resample("W-MON").sum()
    w = w.to_frame("True_Value").reset_index()
    w["weekofyear"] = w["DateTime"].dt.isocalendar().week
    fr_holidays = holidays.CountryHoliday(country)
    w["holiday_week_count"] = w["DateTime"].apply(
        lambda d: sum((d - pd.Timedelta(days=i)).date()
                       in fr_holidays for i in range(7))
    )
    return w

N_LAGS = 7
for i in range(1, N_LAGS + 1):
    weekly[f"lag_{i}"] = weekly["True_Value"].shift(i)
weekly = weekly.dropna()  # -> 200 rows

cut = int(len(weekly) * 0.70)
train, test = weekly.iloc[:cut], weekly.iloc[cut:]
# train: 140 instances, test: 60 instances
\end{lstlisting}
\end{tcolorbox}

\paragraph{XGBoost configuration.} Hyperparameters selected via 200-iteration random search with 4-fold time-series cross-validation (\texttt{TimeSeriesSplit}). Selection criterion: MAE on the validation folds with early stopping (200 rounds).

\begin{table}[H]
\centering
\small
\adjustbox{max width=\columnwidth}{%
\begin{tabular}{ll}
\toprule
\textbf{Parameter} & \textbf{Value} \\
\midrule
\texttt{n\_estimators}           & 1190 \\
\texttt{learning\_rate}          & 0.00497 \\
\texttt{max\_depth}              & 5 \\
\texttt{subsample}               & 0.805 \\
\texttt{colsample\_bytree}       & 0.725 \\
\texttt{min\_child\_weight}      & 6 \\
\texttt{gamma}                   & 0.248 \\
\texttt{reg\_lambda}             & 0.530 \\
\texttt{early\_stopping\_rounds} & 200 \\
\bottomrule
\end{tabular}%
}
\caption{XGBoost hyperparameters.}
\label{tab:xgb-hyperparams}
\end{table}

All predictions are \textbf{one-step-ahead}: each test instance uses true observed values from all preceding weeks as lag features, not recursive model outputs. Test-set performance: $R^2 = 0.69$, MAE $= 20.55$~kWh, RMSE $= 25.04$~kWh.

\paragraph{SHAP computation.} TreeExplainer \citep{lundberg2017unified, lundberg2020local} provides exact Shapley values for tree-based models in polynomial time. For each test instance, the output is a vector of nine SHAP values plus a base value $\phi_0 = 177.37$~kWh (close to the training-set mean of 177.53~kWh), satisfying the efficiency property $\hat{y} = \phi_0 + \sum_{j=1}^{9}\phi_j$.

\begin{tcolorbox}[codebox, title={SHAP Computation}]
\begin{lstlisting}[style=pythonstyle]
import shap
explainer   = shap.TreeExplainer(model)
shap_values = explainer.shap_values(X_test)
base_value  = explainer.expected_value
\end{lstlisting}
\end{tcolorbox}

\paragraph{E5 OOD poisoning.} For E5 (\S\ref{sec:exp5}), each test instance is converted into an out-of-distribution variant by season-dependent lag poisoning. The function below produces the poisoned feature row; XGBoost and TreeExplainer are then re-run on the poisoned inputs to obtain the corresponding prediction and SHAP attributions, and a fresh NLE is generated by the same pipeline as for the original instances (\S\ref{sec:method-nle}).

\begin{tcolorbox}[codebox, title={E5 OOD Poisoning}]
\begin{lstlisting}[style=pythonstyle]
LAG_FEATURES = [f"lag_{i}" for i in range(1, 8)]
rng = np.random.default_rng(RANDOM_SEED)

def poison_instance(row, level="ood"):
    """Create an OOD variant of a test instance.

    Cold-season weeks (44-17): lags pushed
    abnormally LOW (winter-low).
    Warm-season weeks (18-43): lags pushed
    abnormally HIGH (summer-high).
    """
    poisoned = row.copy()
    if level == "baseline":
        return poisoned

    if row["season"] == "cold":
        target = 0.8 * LAG_STATS["min"]    # ~50.8 kWh
    else:
        target = 1.2 * LAG_STATS["max"]    # ~327.9 kWh

    for i, lag in enumerate(LAG_FEATURES):
        decay = 1.0 - i * 0.05             # lag_1=1.0, lag_7=0.7
        noise = rng.uniform(-5, 5)
        poisoned[lag] = target * decay + noise
    return poisoned

# Re-run XGBoost and SHAP on the poisoned inputs
X_ood = poisoned_df[FEATURES].values
poisoned_df["prediction_ood"] = model.predict(X_ood)
shap_ood = shap.TreeExplainer(model).shap_values(X_ood)
\end{lstlisting}
\end{tcolorbox}

\section{NLE Generation Prompt}
\label{sec:appendix-nle-prompt}

The system prompt below is used identically for both generators (zero-shot, $\tau = 1.0$). The human message provides instance data using the template that follows.

\begin{tcolorbox}[systemprompt, title=System Prompt (NLE generation)]
Interpret the time-series forecasting context that follows.

Your goal is to help a non-technical lay user. The user does not have a background in statistical models, machine learning, time-series or explainability methods (e.g.\ SHAP, LIME). The user needs to understand
\begin{itemize}[nosep,leftmargin=*]
\item \textbf{why} the model produced this forecast, and
\item \textbf{how much confidence} can be placed in it.
\end{itemize}

OUTPUT RULES
\begin{itemize}[nosep,leftmargin=*]
\item Write \textbf{up to six bullet points} -- no more.
\item Keep the whole response \textbf{$\leq$ 200 words}.
\item Use plain language, but it is fine to employ key time-series terms such as \textit{lag}, \textit{trend}, \textit{seasonality}, \textit{baseline}, \textit{error}.
\item Do not reveal code.
\end{itemize}
\end{tcolorbox}

\begin{tcolorbox}[humanprompt, title=Human Message Template (NLE generation)]
The following is about time series data with a single-step ahead prediction, where the model predicts the next value in the time series based on previous observations.\\[4pt]
Data Domain: Energy Consumption\\
Dataset Description:\\
\hspace*{1em}- 2{,}075{,}259 measurements from a house in Sceaux, France (near Paris), Dec 2006--Nov 2010.\\
\hspace*{1em}- Recorded at one-minute resolution, then resampled weekly.\\
\hspace*{1em}- Lag\_1..Lag\_7, ISO week number and number of public holidays per week were added as features.\\
\hspace*{1em}- Target: weekly global active power (kWh).\\[4pt]
Model Used: XGBRegressor\\
Model Performance: MAE 20.550, RMSE 25.035, R\textsuperscript{2} 0.686\\[4pt]
Prediction: \{prediction\}\\[4pt]
Instance Features or Context:\\
\hspace*{1em}lag\_1: \{value\}, lag\_2: \{value\}, ..., lag\_7: \{value\},\\
\hspace*{1em}weekofyear: \{value\}, holiday\_week\_count: \{value\}\\[4pt]
SHAP values (sorted by $|$SHAP$|$ descending):\\
\hspace*{1em}\{feature\}: \{shap\_value\}, ...\\
The expected/base value for SHAP: \{base\_value\}
\end{tcolorbox}

\paragraph{Example output.} For the test instance of 2009-11-09 (ISO week 46), XGBoost predicts 175.44~kWh (true value 208.20~kWh; error $-15.7\%$). GPT-4o (zero-shot, $\tau = 1.0$) produces:

\begin{tcolorbox}[promptbox]
\begin{itemize}[nosep,leftmargin=*]
\item \textbf{Why the prediction is 175.44:} The model considers recent weekly consumption from Lag\_1 to Lag\_7 and the time of year. The week of year ($+24.0$) raised the prediction; Lag\_6 ($-13.5$) and Lag\_1 ($-11.8$) lowered it.
\item \textbf{Influences:} Seasonal factors elevated the forecast slightly, while decreases in recent usage (Lag\_6, Lag\_1) weighted the prediction lower.
\item \textbf{Baseline:} The model starts from a baseline of 177.37~kWh and adjusts based on per-feature contributions.
\item \textbf{Reliability:} $R^2 = 0.69$ suggests the model captures about 69\% of the variance, indicating reasonably solid reliability.
\item \textbf{Error margin:} Past MAE = 20.55 and RMSE = 25.03 suggest weekly consumption may differ by this range; moderate uncertainty.
\item \textbf{Confidence:} You can reasonably trust the prediction; consider a $\pm 25$~kWh margin.
\end{itemize}
\end{tcolorbox}

\section{Judge Prompts}
\label{sec:appendix-prompts}

All judge prompts use temperature $\tau = 0.0$ for deterministic evaluation and output structured JSON. The system prompts below are task-specific; the human message that accompanies them contains the instance data and, depending on the condition, the SHAP attributions, the performance metrics, and the NLE. A complete worked example -- system prompt plus human message -- for the E+X+T condition of E1 is shown at the end of this section.

\subsection{E1 and E2: Closeness Task}
\label{sec:appendix-e1-e2-prompt}

The closeness task is shared between E1 (forward simulatability across five information conditions) and E2 (placebic control). The judge system prompt is identical; the human message differs only in which condition's information is included.

\begin{tcolorbox}[systemprompt, title=System Prompt (E1 / E2)]
You are evaluating a machine learning model's prediction accuracy for a time series forecasting task.

You have limited background in machine learning and statistics -- you understand basic concepts but are not an expert.

Your task: Based on all the information provided, assess how large the prediction error is likely to be. Carefully evaluate the features, model context, and any additional information given to form your judgment.

Important: Think carefully about each specific instance. Consider all the details provided -- feature values, patterns, and any explanations. Do not default to a ``safe'' middle-ground answer. The prediction has a specific error magnitude; use the evidence to determine what it is.

Evaluation steps (think through each before giving your final answer):\\
1. Examine the features -- consider the lag values, the time of year (season in France), holidays, and whether the prediction seems reasonable given this context.\\
2. Compare the prediction to recent lag values -- does it follow the recent trend?\\
3. If model performance metrics are provided, use them to estimate a typical error range.\\
4. If SHAP values are provided, consider whether each feature's contribution makes sense given the feature values and context.\\
5. If a natural language explanation is provided, you can use it to better understand the model's reasoning.\\
6. Weigh all available evidence to determine the most likely error magnitude.\\[4pt]
Error bucket definitions (based on absolute percentage error):\\
- small: $[0\%,\,5\%)$\\
- medium: $[5\%,\,15\%)$\\
- large: $[15\%,\,30\%)$\\
- very\_large: $[30\%,\,+\infty)$\\[4pt]
You must respond with EXACTLY this JSON format and nothing else:\\
\{ "error\_bucket": "<small$|$medium$|$large$|$very\_large>",\\
\hspace*{1em} "confidence": <integer 1--5>,\\
\hspace*{1em} "reasoning": "<one sentence>" \}
\end{tcolorbox}

\subsection{E3: Counterfactual Simulatability}
\label{sec:appendix-e3-prompt}

\begin{tcolorbox}[systemprompt, title=System Prompt (E3)]
You are evaluating how a machine learning model's prediction would change if certain input features were modified.

You have limited background in machine learning and statistics -- you understand basic concepts but are not an expert.

Your task: Given the original instance and a proposed feature change, predict whether the model's prediction will be HIGHER, LOWER, or SIMILAR compared to the original prediction.

Evaluation steps:\\
1. Identify which features are being changed and note their current SHAP values.\\
2. For each changed feature, determine the direction of change.\\
3. Consider how the SHAP contribution might change.\\
4. If multiple features are changed, think about their combined effect.\\
5. If a natural language explanation is provided, you can use it to better understand the feature--prediction relationships.\\
6. Estimate the net directional effect and whether it exceeds the 5\% threshold.\\[4pt]
Direction definitions:\\
- higher: prediction increases by more than 5\% relative to the original.\\
- lower: prediction decreases by more than 5\% relative to the original.\\
- similar: prediction changes by less than 5\% in either direction.\\[4pt]
JSON response: \{ "direction": "<higher$|$lower$|$similar>", "confidence": 1--5, "reasoning": "..." \}
\end{tcolorbox}

\subsection{E4: Mental Model Transfer}
\label{sec:appendix-e4-prompt}

\begin{tcolorbox}[systemprompt, title=System Prompt (E4)]
You are evaluating a machine learning model's prediction accuracy for a time series forecasting task.

You have limited background in machine learning and statistics -- you understand basic concepts but are not an expert.

You will first see several example predictions with their KNOWN error outcomes. Study these examples carefully to understand the model's behavior patterns. Then you will evaluate a NEW prediction.

Your task: Based on the patterns you observed in the examples and the information about the new instance, classify the new prediction's likely error bucket.

Error bucket definitions: same as E1/E2.\\[4pt]
JSON response: \{ "error\_bucket": "...", "confidence": 1--5, "reasoning": "..." \}
\end{tcolorbox}

\subsection{E5: Selective Reliance (OOD Detection)}
\label{sec:appendix-e5-prompt}

\begin{tcolorbox}[systemprompt, title=System Prompt (E5)]
You are evaluating a machine learning model's prediction for a time series forecasting task.

You have limited background in machine learning and statistics -- you understand basic concepts but are not an expert.

Your task: Based on all the information provided, decide whether this prediction is reliable or unreliable. Consider whether the input features look normal or unusual, whether they match expected patterns for the time of year, and whether the model can be trusted given these inputs.

Evaluation steps:\\
1. Examine the features -- consider the lag values, the time of year (season in France), holidays, and whether the prediction seems reasonable given this context.\\
2. If SHAP values are provided, consider whether each feature's contribution makes sense given the feature values and context.\\
3. Consider the model performance metrics to gauge general trustworthiness.\\
4. If a natural language explanation is provided, you can use it to better understand the model's reasoning.\\
5. Weigh all evidence to decide whether this prediction can be trusted.\\[4pt]
Reliability:\\
- reliable = inputs look normal and within what the model was trained on.\\
- unreliable = inputs look unusual or far from what the model was trained on.\\[4pt]
JSON response: \{ "reliability": "<reliable$|$unreliable>", "confidence": 1--5, "reasoning": "..." \}
\end{tcolorbox}

\subsection{Worked Judge Human Message (E1, E+X+T)}
\label{sec:appendix-judge-example}

To make the data format the judge actually receives concrete, we reproduce below the complete human message for one judgment: the test instance of 2009-11-09 (ISO week 46) under E1's full E+X+T condition (features + SHAP + metrics + NLE). The same instance is used in the NLE example of Appendix~\ref{sec:appendix-nle-prompt}; the NLE text below is the GPT-4o output shown there. Other conditions strip the corresponding blocks (e.g., the X condition removes the Metrics block and the NLE block); other experiments substitute their own task framing while retaining the same data-block layout.

\begin{tcolorbox}[humanprompt, title={Human Message Example (E1, E+X+T condition)}]
The following is about time series data with a single-step ahead prediction, where the model predicts the next value in the time series based on previous observations.\\[4pt]
Data Domain: Energy Consumption\\
Dataset Description:\\
\hspace*{1em}-- 2{,}075{,}259 measurements from a house in Sceaux, France (near Paris), Dec 2006--Nov 2010.\\
\hspace*{1em}-- Recorded at one-minute resolution, then resampled weekly.\\
\hspace*{1em}-- Lag\_1..Lag\_7, ISO week number and number of public holidays per week were added as features.\\
\hspace*{1em}-- Target: weekly global active power (kWh).\\[4pt]
Model Used: XGBRegressor\\
Model Performance:\\
\hspace*{1em}-- MAE: 20.550\\
\hspace*{1em}-- RMSE: 25.035\\
\hspace*{1em}-- R\textsuperscript{2}: 0.686\\[4pt]
Prediction: 175.44\\[4pt]
Instance Features or Context:\\
\hspace*{1em}lag\_1: 184.727, lag\_2: 216.379, lag\_3: 197.242, lag\_4: 191.376,\\
\hspace*{1em}lag\_5: 166.269, lag\_6: 162.277, lag\_7: 189.772,\\
\hspace*{1em}weekofyear: 46, holiday\_week\_count: 0\\[4pt]
SHAP values (sorted by $|$magnitude$|$, descending):\\
\hspace*{1em}weekofyear: $+$24.032\\
\hspace*{1em}lag\_6: $-$13.501\\
\hspace*{1em}lag\_1: $-$11.754\\
\hspace*{1em}lag\_2: $+$3.363\\
\hspace*{1em}lag\_7: $-$3.246\\
\hspace*{1em}lag\_4: $-$2.308\\
\hspace*{1em}lag\_5: $+$1.771\\
\hspace*{1em}lag\_3: $-$0.263\\
\hspace*{1em}holiday\_week\_count: $-$0.019\\
The expected/base value for SHAP: 177.366\\[4pt]
Natural language explanation of this prediction:\\
-- Why the prediction is 175.44: The model considers recent weekly consumption from Lag\_1 to Lag\_7 and the time of year. The week of year ($+$24.0) raised the prediction; Lag\_6 ($-$13.5) and Lag\_1 ($-$11.8) lowered it.\\
-- Influences: Seasonal factors elevated the forecast slightly, while decreases in recent usage (Lag\_6, Lag\_1) weighted the prediction lower.\\
-- Baseline: The model starts from a baseline of 177.37~kWh and adjusts based on per-feature contributions.\\
-- Reliability: R\textsuperscript{2}$=0.69$ suggests the model captures about 69\% of the variance, indicating reasonably solid reliability.\\
-- Error margin: Past MAE $=20.55$ and RMSE $=25.03$ suggest weekly consumption may differ by this range; moderate uncertainty.\\
-- Confidence: You can reasonably trust the prediction; consider a $\pm 25$~kWh margin.\\[4pt]
Based on all the information above, classify this prediction's likely error bucket.
\end{tcolorbox}

\noindent The true weekly consumption for this instance was 208.20~kWh, giving an absolute percentage error of 15.7\% (the \emph{large} bucket).

\section{Full Statistical Results}
\label{sec:appendix-stats}

\paragraph{Modelling framework.} All experiments share a within-instance repeated-measures design. Accuracy: GLMM (binomial, logit link) via \texttt{lme4} \citep{bates2015lme4}, fixed effects condition $+$ judge ($+$ interaction where convergence allows), random intercept by instance. Optimiser: bobyqa. Confidence: CLMM (cumulative logit) via \texttt{ordinal} \citep{christensen2019ordinal}. Overconfidence: CLMM with correctness $\times$ condition interaction. Omnibus tests: likelihood-ratio tests (LRT) on nested models. Planned contrasts: estimated marginal means with Holm correction \citep{lenth2024emmeans}. Effect size: odds ratios (OR) with 95\% CIs. Bayesian analyses: \texttt{brms} \citep{burkner2017brms}, $\mathcal{N}(0,\,1.5)$ priors on fixed effects \citep{gelman2008weakly}, half-$t_3(0,\,2.5)$ on random-intercept SD, 4 chains $\times$ 4000 iterations. ROPE on log-odds scale: $[-0.18,\,+0.18]$ ($\text{OR} \in [0.84,\,1.20]$).

This appendix reports the full statistical output for the Part~2 analyses described in \S\ref{sec:method-stats}. Each experiment is presented with a common structure: (1)~descriptives, (2)~GLMM fixed effects and model fit, (3)~CLMM summary for confidence, (4)~calibration analysis (mixed-effects model, Somers' $D$, empirical calibration, conditional confidence), (5)~model diagnostics, (6)~Bayesian ROPE, and (7)~sensitivity analyses (judge-specific, generator, same-family, random slopes, and experiment-specific covariates). Where the model structure permits, estimated marginal means and a planned-contrasts / pairwise-comparisons table are also reported -- specifically, for experiments with three or more conditions (E1, E2). Experiments with only two conditions (E3, E4) or a $2\times 2$ interaction design (E5) do not require a separate pairwise table because the fixed-effect coefficient in the primary GLMM is itself the pairwise contrast (or the interaction term is the relevant simple-effects decomposition). All primary models use the interaction specification of the primary GLMM.

\paragraph{Reading the diagnostics line.} Each experiment reports a one-line diagnostics summary of the form ``Convergence: OK. Singularity: OK. DHARMa uniformity $p = \cdot$. Dispersion $p = \cdot$.'' The four checks mean: (i)~the maximum-likelihood optimisation converged without warnings; (ii)~the random-intercept variance is not estimated on the boundary (non-zero, not a singular fit); (iii)~the DHARMa simulated-residual distribution is uniform (a large $p$-value is \emph{good} -- it indicates no evidence of miscalibration); and (iv)~the dispersion test detects no over- or under-dispersion of the binomial residuals (again, large $p = $ no problem).

\paragraph{Reading ``NA'' rows in empirical calibration tables.} Some tables include an ``NA'' row under \emph{Confidence level}. These are judgments where the confidence rating could not be parsed from the judge's response (e.g., the judge wrote a response that did not match the expected 1--5 integer format). All such judgments are retained in the primary model (via \texttt{confidence\_num} treated as continuous when the response parses, and treated as missing when it does not) but tabulated separately so their accuracy rate -- which is essentially zero because these responses also typically fail to commit to a valid outcome bucket -- is visible rather than hidden.

\subsection{Experiment 1: Forward Simulatability}
\label{app:p2-e1}

\paragraph*{Descriptives}

\begin{table}[H]
\centering\small
\adjustbox{max width=\columnwidth}{%
\begin{tabular}{lcccc}
\toprule
\textbf{Condition} & \textbf{$N$} & \textbf{Accuracy} & \textbf{Mean conf.} & \textbf{Median conf.} \\
\midrule
Baseline & 120 & .358 & 3.33 & 3 \\
X        & 120 & .383 & 3.41 & 3 \\
T        & 120 & .392 & 3.50 & 4 \\
X+T      & 120 & .417 & 3.81 & 4 \\
E+X+T    & 240 & .392 & 3.92 & 4 \\
\bottomrule
\end{tabular}%
}
\caption[E1 descriptive statistics by condition.]{E1 descriptives. True bucket distribution across the 60 test instances: small 12, medium 29, large 16, very-large 3 (192, 348, 144, 36 judgments respectively).}
\label{tab:app-e1-desc}
\end{table}

\paragraph*{GLMM Fixed Effects (Accuracy)}

\begin{table}[H]
\centering\small
\resizebox{\columnwidth}{!}{%
\begin{tabular}{lrrrr}
\toprule
\textbf{Effect} & \textbf{Estimate} & \textbf{SE} & \textbf{OR} & \textbf{$p$} \\
\midrule
Intercept                  & $-$0.656 & 0.456 & 0.519 & .151 \\
X (SHAP)                   & $-$0.243 & 0.478 & 0.785 & .612 \\
T (Metrics)                & $+$0.617 & 0.481 & 1.853 & .200 \\
X+T                        & $+$0.744 & 0.483 & 2.105 & .123 \\
E+X+T (NLE)                & $+$0.367 & 0.414 & 1.444 & .375 \\
Judge: GPT-4o              & $-$1.114 & 0.492 & 0.329 & .024 \\
X $\times$ Judge           & $+$0.871 & 0.686 & 2.389 & .205 \\
T $\times$ Judge           & $-$0.749 & 0.697 & 0.473 & .283 \\
X+T $\times$ Judge         & $-$0.615 & 0.694 & 0.541 & .375 \\
E+X+T $\times$ Judge       & $-$0.238 & 0.599 & 0.788 & .691 \\
\midrule
Random intercept SD        & \multicolumn{4}{c}{2.291 (variance 5.248; 60 instance groups)} \\
Fit                        & \multicolumn{4}{c}{AIC $=$ 731.0; BIC $=$ 781.4; $\log L = -354.5$} \\
\bottomrule
\end{tabular}%
}
\caption[E1 GLMM fixed effects.]{E1 GLMM fixed effects. Reference: Baseline, DeepSeek-R1. Omnibus LRT: $\chi^2(8) = 8.23$, $p = .41$. $N = 720$, 60 instance groups. No judge~$\times$~condition interaction is significant.}
\label{tab:app-e1-glmm}
\end{table}

\paragraph*{Estimated Marginal Means (Accuracy)}

\begin{table}[H]
\centering\small
\adjustbox{max width=\columnwidth}{%
\begin{tabular}{lcccc}
\toprule
\textbf{Condition} & \textbf{Prob.} & \textbf{SE} & \textbf{95\% CI} & \\
\midrule
\multicolumn{4}{l}{\textit{Marginal (averaged over judges):}} \\
Baseline & .229 & .070 & [.120, .392] & \\
X        & .265 & .076 & [.143, .437] & \\
T        & .275 & .079 & [.149, .451] & \\
X+T      & .315 & .085 & [.176, .498] & \\
E+X+T    & .276 & .071 & [.160, .432] & \\
\midrule
\multicolumn{4}{l}{\textit{DeepSeek-R1:}} \\
Baseline & .342 & .103 & [.175, .559] & \\
X        & .289 & .094 & [.142, .500] & \\
T        & .490 & .114 & [.282, .702] & \\
X+T      & .522 & .114 & [.308, .728] & \\
E+X+T    & .428 & .095 & [.260, .616] & \\
\midrule
\multicolumn{4}{l}{\textit{GPT-4o:}} \\
Baseline & .146 & .059 & [.063, .302] & \\
X        & .242 & .085 & [.115, .441] & \\
T        & .130 & .054 & [.055, .277] & \\
X+T      & .162 & .064 & [.072, .328] & \\
E+X+T    & .162 & .055 & [.081, .299] & \\
\bottomrule
\end{tabular}%
}
\caption[E1 estimated marginal means.]{E1 EMMs on the probability scale, marginal and by judge. Full 5-by-3 grid (no selected subset).}
\label{tab:app-e1-emm}
\end{table}

\paragraph*{Planned Contrasts and All Pairwise}

\begin{table}[H]
\centering\small
\adjustbox{max width=\columnwidth}{%
\begin{tabular}{lcccc}
\toprule
\textbf{Contrast} & \textbf{OR} & \textbf{SE} & $z$ & \textbf{$p$} \\
\midrule
\multicolumn{5}{l}{\textit{Planned:}} \\
E+X+T vs X+T (NLE marginal effect)   & 0.83 & 0.25 & $-$0.63 & .530 \\
X+T vs Baseline (information effect) & 1.55 & 0.54 & $+$1.26 & .208 \\
\midrule
\multicolumn{5}{l}{\textit{All pairwise (Holm-corrected, all $p_{\text{Holm}} = 1.0$):}} \\
Baseline vs X     & 0.825 & 0.283 & $-$0.56 & 1.0 \\
Baseline vs T     & 0.785 & 0.273 & $-$0.70 & 1.0 \\
Baseline vs X+T   & 0.646 & 0.224 & $-$1.26 & 1.0 \\
Baseline vs E+X+T & 0.780 & 0.234 & $-$0.83 & 1.0 \\
X vs T            & 0.951 & 0.328 & $-$0.15 & 1.0 \\
X vs X+T          & 0.783 & 0.269 & $-$0.71 & 1.0 \\
X vs E+X+T        & 0.946 & 0.279 & $-$0.19 & 1.0 \\
T vs X+T          & 0.823 & 0.287 & $-$0.56 & 1.0 \\
T vs E+X+T        & 0.994 & 0.300 & $-$0.02 & 1.0 \\
X+T vs E+X+T      & 1.207 & 0.362 & $+$0.63 & 1.0 \\
\bottomrule
\end{tabular}%
}
\caption[E1 planned contrasts and all pairwise comparisons.]{E1 condition contrasts on accuracy. No pair survives Holm correction.}
\label{tab:app-e1-pairs}
\end{table}

\paragraph*{CLMM (Confidence)}

\begin{table}[H]
\centering\small
\adjustbox{max width=\columnwidth}{%
\begin{tabular}{lrrrr}
\toprule
\textbf{Effect} & \textbf{Estimate} & \textbf{SE} & \textbf{OR} & \textbf{$p$} \\
\midrule
X                  & $+$0.267 & 0.472 & 1.306 & .571 \\
T                  & $+$2.097 & 0.456 & 8.14  & $<$.001 \\
X+T                & $+$2.496 & 0.471 & 12.13 & $<$.001 \\
E+X+T              & $+$3.705 & 0.465 & 40.64 & $<$.001 \\
Judge: GPT-4o      & $+$2.172 & 0.469 & 8.78  & $<$.001 \\
X $\times$ Judge      & $+$0.274 & 0.640 & 1.32  & .669 \\
T $\times$ Judge      & $-$2.723 & 0.637 & 0.066 & $<$.001 \\
X+T $\times$ Judge    & $+$1.637 & 1.143 & 5.14  & .152 \\
E+X+T $\times$ Judge  & $+$0.471 & 0.978 & 1.60  & .630 \\
\midrule
Threshold $\theta_{2|3}$ & $-$3.264 & 0.496 & & \\
Threshold $\theta_{3|4}$ & $+$1.819 & 0.374 & & \\
Threshold $\theta_{4|5}$ & $+$10.871 & 1.062 & & \\
Random intercept SD       & \multicolumn{4}{c}{0.665 (60 groups)} \\
\bottomrule
\end{tabular}%
}
\caption[E1 CLMM (confidence) fixed effects and thresholds.]{E1 CLMM on confidence ($N = 625$ non-missing). Reference: Baseline, DeepSeek-R1. Huge positive shifts for T, X+T, E+X+T indicate information-scaled confidence inflation.}
\label{tab:app-e1-clmm}
\end{table}

\begin{table}[H]
\centering\small
\adjustbox{max width=\columnwidth}{%
\begin{tabular}{lcccc}
\toprule
\textbf{Contrast (log-odds)} & \textbf{Estimate} & \textbf{SE} & $z$ & \textbf{$p$ (Holm)} \\
\midrule
Baseline vs X          & $-$0.404 & 0.320 & $-$1.26 & .623 \\
Baseline vs T          & $-$0.736 & 0.315 & $-$2.34 & .078 \\
Baseline vs X+T        & $-$3.314 & 0.577 & $-$5.74 & $<$.0001 \\
Baseline vs E+X+T      & $-$3.940 & 0.503 & $-$7.84 & $<$.0001 \\
X vs T                 & $-$0.332 & 0.312 & $-$1.06 & .623 \\
X vs X+T               & $-$2.911 & 0.574 & $-$5.07 & $<$.0001 \\
X vs E+X+T             & $-$3.537 & 0.497 & $-$7.12 & $<$.0001 \\
T vs X+T               & $-$2.579 & 0.567 & $-$4.55 & $<$.0001 \\
T vs E+X+T             & $-$3.205 & 0.489 & $-$6.56 & $<$.0001 \\
X+T vs E+X+T           & $-$0.626 & 0.587 & $-$1.07 & .623 \\
\bottomrule
\end{tabular}%
}
\caption[E1 confidence pairwise comparisons.]{E1 confidence pairwise comparisons on the log-odds scale (Holm-corrected). Eight of ten pairs differ at $p < .0001$. The information-bearing conditions (X+T, E+X+T) sit far above Baseline/X/T; the NLE does not add a further significant increment over X+T (last row, $p = .62$).}
\label{tab:app-e1-conf-pairs}
\end{table}

\paragraph*{Calibration}

\begin{table}[H]
\centering\small
\adjustbox{max width=\columnwidth}{%
\begin{tabular}{lrrrr}
\toprule
\textbf{Effect} & \textbf{Estimate} & \textbf{SE} & $z$ & \textbf{$p$} \\
\midrule
Intercept                    & $+$4.350 & 1.766 & $+$2.46 & .014 \\
confidence\_num              & $-$1.519 & 0.522 & $-$2.91 & .004 \\
X                            & $-$2.965 & 2.492 & $-$1.19 & .234 \\
T                            & $-$3.058 & 2.640 & $-$1.16 & .247 \\
X+T                          & $-$3.168 & 2.976 & $-$1.06 & .287 \\
E+X+T                        & $-$7.416 & 3.169 & $-$2.34 & .019 \\
Judge: GPT-4o                & $-$0.678 & 0.240 & $-$2.82 & .005 \\
conf $\times$ X              & $+$0.963 & 0.727 & $+$1.32 & .186 \\
conf $\times$ T              & $+$1.098 & 0.755 & $+$1.45 & .146 \\
conf $\times$ X+T            & $+$1.155 & 0.812 & $+$1.42 & .155 \\
\textbf{conf $\times$ E+X+T} & $\mathbf{+2.260}$ & \textbf{0.849} & $\mathbf{+2.66}$ & $\mathbf{.008}$ \\
\midrule
Random intercept SD          & \multicolumn{4}{c}{2.49} \\
\bottomrule
\end{tabular}%
}
\caption[E1 calibration model coefficients.]{E1 calibration model: $\text{correct} \sim \text{confidence} \times \text{condition} + \text{judge} + (1 \mid \text{instance})$. The confidence $\times$ E+X+T interaction ($p = .008$) is the only significant calibration slope difference: in the full-NLE condition, higher confidence tracks accuracy better than at Baseline. The main confidence coefficient is the reference (Baseline) slope.}
\label{tab:app-e1-cal}
\end{table}

\begin{table}[H]
\centering\small
\adjustbox{max width=\columnwidth}{%
\begin{tabular}{lcc}
\toprule
\textbf{Condition} & \textbf{Somers' $D$} & $N$ \\
\midrule
Baseline & $-$0.307 & 120 \\
X        & $-$0.177 & 120 \\
T        & $+$0.050 & 120 \\
X+T      & $-$0.028 & 120 \\
E+X+T    & $+$0.037 & 240 \\
\bottomrule
\end{tabular}%
}
\caption[E1 Somers' $D$ by condition.]{E1 Somers' $D$ (confidence--accuracy ordinal association). Negative $D$ indicates that higher stated confidence is associated with \emph{lower} accuracy. Baseline shows the worst calibration; the full-NLE condition is the only one with (very weakly) positive calibration.}
\label{tab:app-e1-somers}
\end{table}

\begin{table}[H]
\centering\small
\adjustbox{max width=\columnwidth}{%
\begin{tabular}{llcc}
\toprule
\textbf{Condition} & \textbf{Confidence level} & \textbf{Accuracy} & $n$ \\
\midrule
Baseline & 2  & .400 & 5 \\
         & 3  & .532 & 62 \\
         & 4  & .195 & 41 \\
         & NA & .000 & 12 \\
X        & 2  & 1.00 & 1 \\
         & 3  & .492 & 61 \\
         & 4  & .333 & 45 \\
         & NA & .000 & 13 \\
T        & 3  & .440 & 50 \\
         & 4  & .490 & 51 \\
         & NA & .000 & 19 \\
X+T      & 3  & .524 & 21 \\
         & 4  & .447 & 85 \\
         & 5  & 1.00 & 1 \\
         & NA & .000 & 13 \\
E+X+T    & 3  & .389 & 18 \\
         & 4  & .470 & 183 \\
         & 5  & 1.00 & 1 \\
         & NA & .000 & 38 \\
\bottomrule
\end{tabular}%
}
\caption[E1 empirical calibration.]{E1 empirical calibration: observed accuracy at each stated confidence level, per condition. A perfectly calibrated rater shows monotonically increasing accuracy with confidence; in Baseline the pattern is inverted (confidence 4 has lower accuracy than confidence 3).}
\label{tab:app-e1-calcurve}
\end{table}

\begin{table}[H]
\centering\small
\adjustbox{max width=\columnwidth}{%
\begin{tabular}{lcccc}
\toprule
\textbf{Condition} & \textbf{Mean conf.\ | incorrect} & $n$ & \textbf{Mean conf.\ | correct} & $n$ \\
\midrule
Baseline & 3.462 & 77  & 3.140 & 43 \\
X        & 3.492 & 74  & 3.304 & 46 \\
T        & 3.481 & 73  & 3.532 & 47 \\
X+T      & 3.825 & 70  & 3.800 & 50 \\
E+X+T    & 3.898 & 146 & 3.936 & 94 \\
\bottomrule
\end{tabular}%
}
\caption[E1 conditional confidence.]{E1 mean confidence split by correctness. Confidence when wrong rises from 3.46 (Baseline) to 3.90 (E+X+T) -- wrong answers become more confident with more information, the definition of overconfidence.}
\label{tab:app-e1-condconf}
\end{table}

\paragraph*{Diagnostics}

Convergence: OK. Singularity: OK. DHARMa uniformity $p = .179$. Dispersion $p = .640$. No violations.

\paragraph*{Bayesian ROPE}

\begin{table}[H]
\centering\small
\adjustbox{max width=\columnwidth}{%
\begin{tabular}{lcccc}
\toprule
\textbf{Parameter} & \textbf{Post.\ OR} & \textbf{95\% CrI} & \textbf{ROPE \%} \\
\midrule
Intercept     & 0.580 & [0.252, 1.291] & -- \\
Judge: GPT-4o & 0.284 & [0.187, 0.430] & -- \\
X             & 1.174 & [0.607, 2.251] & 37.5 \\
T             & 1.241 & [0.641, 2.401] & 33.6 \\
X+T           & 1.491 & [0.751, 2.913] & 21.1 \\
E+X+T         & 1.254 & [0.703, 2.226] & 34.8 \\
\bottomrule
\end{tabular}%
}
\caption[E1 Bayesian posterior ORs and ROPE analysis.]{E1 Bayesian posterior ORs for all fixed effects. ROPE $= [-0.18, +0.18]$ on log-odds $\Leftrightarrow \text{OR} \in [0.84, 1.20]$. No condition reaches the 95\% ROPE threshold, but none excludes 1 either -- the posteriors are consistent with small, non-significant effects.}
\label{tab:app-e1-rope}
\end{table}

\paragraph*{Sensitivity}

\paragraph{Judge-specific.} Stats-plan item: refit the primary GLMM separately on the GPT-4o and DeepSeek-R1 subsets to verify that the marginal effect is stable across raters rather than driven by one judge. Per-judge model: $\text{correct} \sim \text{condition} + (1 \mid \text{instance})$.

\begin{table}[H]
\centering\small
\adjustbox{max width=\columnwidth}{%
\begin{tabular}{lccccc}
\toprule
\textbf{Judge} & Intercept & X & T & X+T & E+X+T \\
\midrule
DeepSeek-R1 & 0.313 & 0.653 & 2.927 & 3.649 & 1.896 \\
GPT-4o      & 0.252 & 1.739 & 0.888 & 1.123 & 1.123 \\
\bottomrule
\end{tabular}%
}
\caption[E1 judge-specific ORs.]{E1 judge-specific ORs (from separate GLMMs per judge). Neither judge's condition effects reach significance; they disagree in direction only for T and X+T, consistent with the non-significant judge~$\times$~condition interactions in the primary GLMM (Table~\ref{tab:app-e1-glmm}).}
\label{tab:app-e1-judge}
\end{table}

\paragraph{Generator and same-family.} Fitted on NLE rows only ($N = 240$): intercept $-0.350$ (SE $0.478$, $p = .46$); \texttt{generator$_{\text{GPT-4o}}$} $+0.282$ (SE $0.377$, $p = .455$); judge effect $-1.514$ (SE $0.413$, $p = .0002$). Same-family bias: $\text{OR} = 1.00$, $p = 1.0$. Which LLM wrote the NLE does not affect accuracy; the dominant structured effect is judge identity.

\paragraph{Random slopes.} The model $\text{correct} \sim \text{condition} \times \text{judge} + (\text{condition} \mid \text{instance})$ is singular, confirming that the intercept-only random structure is sufficient ().

\subsection{Experiment 2: Placebic Control}
\label{app:p2-e2}

\paragraph*{Descriptives}

\begin{table}[H]
\centering\small
\adjustbox{max width=\columnwidth}{%
\begin{tabular}{lcccc}
\toprule
\textbf{Condition} & \textbf{$N$} & \textbf{Accuracy} & \textbf{Mean conf.} & \textbf{Median conf.} \\
\midrule
Baseline    & 120 & .450 & 3.81 & 4 \\
Real NLE    & 240 & .479 & 3.95 & 4 \\
Placebo NLE & 240 & .458 & 3.90 & 4 \\
\bottomrule
\end{tabular}%
}
\caption[E2 descriptive statistics by condition.]{E2 descriptives. By judge: DeepSeek-R1 -- Baseline .533/3.62, Real .525/3.90, Placebo .517/3.71; GPT-4o -- Baseline .367/4.00, Real .433/4.00, Placebo .400/4.09.}
\label{tab:app-e2-desc}
\end{table}

\paragraph*{GLMM Fixed Effects (Accuracy)}

\begin{table}[H]
\centering\small
\resizebox{\columnwidth}{!}{%
\begin{tabular}{lrrrr}
\toprule
\textbf{Effect} & \textbf{Estimate} & \textbf{SE} & \textbf{OR} & \textbf{$p$} \\
\midrule
Intercept                      & $+$0.134 & 0.470 & 1.144 & .775 \\
Real NLE                       & $-$0.067 & 0.425 & 0.935 & .875 \\
Placebo NLE                    & $-$0.133 & 0.425 & 0.875 & .754 \\
Judge: GPT-4o                  & $-$1.302 & 0.497 & 0.272 & .009 \\
Real NLE $\times$ Judge        & $+$0.584 & 0.602 & 1.793 & .332 \\
Placebo NLE $\times$ Judge     & $+$0.393 & 0.601 & 1.481 & .514 \\
\midrule
Random intercept SD            & \multicolumn{4}{c}{2.400 (variance 5.76; 60 groups)} \\
Fit                            & \multicolumn{4}{c}{AIC $=$ 609.2; BIC $=$ 640.0; $\log L = -297.6$} \\
\bottomrule
\end{tabular}%
}
\caption[E2 GLMM fixed effects.]{E2 GLMM fixed effects. Reference: Baseline, DeepSeek-R1. Omnibus LRT: $\chi^2(4) = 1.55$, $p = .82$. $N = 600$, 60 instance groups. No interaction significant.}
\label{tab:app-e2-glmm}
\end{table}

\paragraph*{Estimated Marginal Means (Accuracy)}

\begin{table}[H]
\centering\small
\adjustbox{max width=\columnwidth}{%
\begin{tabular}{lcccc}
\toprule
\textbf{Condition} & \textbf{Prob.} & \textbf{SE} & \textbf{95\% CI} & \\
\midrule
\multicolumn{4}{l}{\textit{Marginal:}} \\
Baseline    & .374 & .094 & [.213, .568] & \\
Real NLE    & .428 & .089 & [.269, .603] & \\
Placebo NLE & .388 & .086 & [.238, .564] & \\
\midrule
\multicolumn{4}{l}{\textit{DeepSeek-R1:}} \\
Baseline    & .534 & .117 & [.313, .742] & \\
Real NLE    & .517 & .100 & [.328, .701] & \\
Placebo NLE & .500 & .100 & [.313, .687] & \\
\midrule
\multicolumn{4}{l}{\textit{GPT-4o:}} \\
Baseline    & .237 & .086 & [.109, .442] & \\
Real NLE    & .343 & .091 & [.192, .534] & \\
Placebo NLE & .287 & .083 & [.154, .471] & \\
\bottomrule
\end{tabular}%
}
\caption[E2 estimated marginal means.]{E2 EMMs on the probability scale, marginal and by judge.}
\label{tab:app-e2-emm}
\end{table}

\paragraph*{Planned Contrasts and All Pairwise}

\begin{table}[H]
\centering\small
\adjustbox{max width=\columnwidth}{%
\begin{tabular}{lcccc}
\toprule
\textbf{Contrast} & \textbf{OR} & \textbf{SE} & $z$ & \textbf{$p$ (Holm)} \\
\midrule
Real vs Baseline         & 1.25 & 0.38 & $+$0.75 & 1.0 \\
Placebo vs Baseline      & 1.07 & 0.32 & $+$0.21 & 1.0 \\
Real vs Placebo          & 1.18 & 0.29 & $+$0.66 & 1.0 \\
\bottomrule
\end{tabular}%
}
\caption[E2 condition contrasts.]{E2 planned and pairwise condition contrasts (Holm-corrected). The critical Real-vs-Placebo contrast ($p = 1.0$) shows that real and placebo NLEs produce indistinguishable accuracy; see also the Bayesian equivalence analysis below.}
\label{tab:app-e2-pairs}
\end{table}

\paragraph*{CLMM (Confidence)}

The interaction CLMM produces NaN standard errors because the random-intercept variance for confidence collapses to zero (confidence does not vary meaningfully between instances). Fixed-effect point estimates are still informative. The additive model is used as the primary fit.

\begin{table}[H]
\centering\small
\adjustbox{max width=\columnwidth}{%
\begin{tabular}{lr}
\toprule
\textbf{Effect} & \textbf{Estimate} \\
\midrule
\multicolumn{2}{l}{\textit{Interaction CLMM (NaN SEs on interactions):}} \\
Real NLE                 & $+$1.656 \\
Placebo NLE              & $+$0.409 \\
Judge: GPT-4o            & $+$3.218 \\
Real NLE $\times$ Judge  & $-$1.656 \\
Placebo NLE $\times$ Judge & $+$1.083 \\
\midrule
\multicolumn{2}{l}{\textit{Thresholds:}} \\
$\theta_{3|4}$           & $-$0.473 \\
$\theta_{4|5}$           & $+$6.909 \\
\bottomrule
\end{tabular}%
}
\caption[E2 CLMM fixed effects.]{E2 CLMM (confidence). SEs suppressed because the random-intercept variance estimate is essentially zero; the additive model below is used for significance testing.}
\label{tab:app-e2-clmm}
\end{table}

\begin{table}[H]
\centering\small
\adjustbox{max width=\columnwidth}{%
\begin{tabular}{lrrr}
\toprule
\textbf{Contrast (additive CLMM)} & \textbf{Estimate} & \textbf{SE} & \textbf{$p$ (Holm)} \\
\midrule
\textbf{Baseline vs Real NLE}      & $\mathbf{-1.101}$ & \textbf{0.317} & $\mathbf{.0015}$ \\
Baseline vs Placebo NLE            & $-$0.683 & 0.305 & .0506 \\
Real NLE vs Placebo NLE            & $+$0.419 & 0.272 & .124 \\
\bottomrule
\end{tabular}%
}
\caption[E2 confidence pairwise contrasts.]{E2 confidence pairwise contrasts from the additive CLMM. Real NLEs significantly elevate confidence relative to Baseline; placebo NLEs do so marginally; real and placebo are not significantly distinguishable, consistent with the presence-not-content hypothesis.}
\label{tab:app-e2-conf}
\end{table}

\paragraph*{Calibration}

\begin{table}[H]
\centering\small
\adjustbox{max width=\columnwidth}{%
\begin{tabular}{lrrrr}
\toprule
\textbf{Effect} & \textbf{Estimate} & \textbf{SE} & $z$ & \textbf{$p$} \\
\midrule
Intercept                  & $+$1.174 & 2.475 & $+$0.47 & .635 \\
confidence\_num            & $-$0.335 & 0.650 & $-$0.52 & .607 \\
Real NLE                   & $+$1.175 & 3.705 & $+$0.32 & .751 \\
Placebo NLE                & $-$0.878 & 2.842 & $-$0.31 & .757 \\
\textbf{Judge: GPT-4o}     & $\mathbf{-0.854}$ & \textbf{0.240} & $\mathbf{-3.56}$ & $\mathbf{.0004}$ \\
conf $\times$ Real NLE     & $-$0.225 & 0.950 & $-$0.24 & .813 \\
conf $\times$ Placebo NLE  & $+$0.251 & 0.740 & $+$0.34 & .735 \\
\midrule
Random intercept SD        & \multicolumn{4}{c}{2.41} \\
\bottomrule
\end{tabular}%
}
\caption[E2 calibration model coefficients.]{E2 calibration model: $\text{correct} \sim \text{confidence} \times \text{condition} + \text{judge} + (1 \mid \text{instance})$. Neither confidence $\times$ Real nor confidence $\times$ Placebo interaction is significant: the confidence--accuracy mapping does not differ between real and placebo NLEs.}
\label{tab:app-e2-cal}
\end{table}

\begin{table}[H]
\centering\small
\adjustbox{max width=\columnwidth}{%
\begin{tabular}{lcc}
\toprule
\textbf{Condition} & \textbf{Somers' $D$} & $N$ \\
\midrule
Baseline    & $-$0.056 & 120 \\
Real NLE    & $-$0.004 & 240 \\
Placebo NLE & $-$0.023 & 240 \\
\bottomrule
\end{tabular}%
}
\caption[E2 Somers' $D$ by condition.]{E2 Somers' $D$. All near zero -- poor calibration throughout, with no difference between real and placebo.}
\label{tab:app-e2-somers}
\end{table}

\begin{table}[H]
\centering\small
\adjustbox{max width=\columnwidth}{%
\begin{tabular}{llcc}
\toprule
\textbf{Condition} & \textbf{Confidence level} & \textbf{Accuracy} & $n$ \\
\midrule
Baseline    & 3 & .522 & 23 \\
            & 4 & .433 & 97 \\
Real NLE    & 3 & .500 & 14 \\
            & 4 & .478 & 224 \\
            & 5 & .500 & 2 \\
Placebo NLE & 3 & .385 & 39 \\
            & 4 & .500 & 186 \\
            & 5 & .133 & 15 \\
\bottomrule
\end{tabular}%
}
\caption[E2 empirical calibration.]{E2 empirical calibration: accuracy at each stated confidence level, per condition. Neither real nor placebo NLEs produce a monotonically increasing calibration curve.}
\label{tab:app-e2-calcurve}
\end{table}

\begin{table}[H]
\centering\small
\adjustbox{max width=\columnwidth}{%
\begin{tabular}{lcccc}
\toprule
\textbf{Condition} & \textbf{Conf.\ | incorrect} & $n$ & \textbf{Conf.\ | correct} & $n$ \\
\midrule
Baseline    & 3.833 &  66 & 3.778 &  54 \\
Real NLE    & 3.952 & 125 & 3.948 & 115 \\
Placebo NLE & 3.915 & 130 & 3.882 & 110 \\
\bottomrule
\end{tabular}%
}
\caption[E2 conditional confidence.]{E2 mean confidence split by correctness. Confidence is essentially identical for correct and incorrect judgments within every condition -- confidence is decoupled from accuracy throughout.}
\label{tab:app-e2-condconf}
\end{table}

\paragraph*{Diagnostics}

Convergence: OK. Singularity: OK. DHARMa uniformity $p = .605$. Dispersion $p = .496$. No violations.

\paragraph*{Bayesian ROPE and Real-vs-Placebo Equivalence}

\begin{table}[H]
\centering\small
\adjustbox{max width=\columnwidth}{%
\begin{tabular}{lccc}
\toprule
\textbf{Parameter} & \textbf{Post.\ OR} & \textbf{95\% CrI} & \textbf{ROPE \%} \\
\midrule
Intercept               & 0.949 & [0.395, 2.208] & -- \\
Judge: GPT-4o           & 0.402 & [0.255, 0.624] & -- \\
Real NLE                & 1.255 & [0.690, 2.270] & 34.7 \\
Placebo NLE             & 1.061 & [0.592, 1.933] & 44.7 \\
\midrule
\textbf{Real} $-$ \textbf{Placebo} & \textbf{1.185} & \textbf{[0.714, 1.938]} & \textbf{43.1} \\
\bottomrule
\end{tabular}%
}
\caption[E2 Bayesian ROPE and Real-vs-Placebo equivalence.]{E2 Bayesian ROPE analysis. Both NLE conditions have posterior ORs close to 1 with wide CrIs that include 1. The direct Real-vs-Placebo equivalence test yields 43.1\% inside ROPE: insufficient for formal practical equivalence (${\geq}95\%$), but consistent with no meaningful difference between real and placebo NLEs.}
\label{tab:app-e2-rope}
\end{table}

\paragraph*{Sensitivity}

\paragraph{Judge-specific.} Stats-plan item: refit the primary GLMM separately per judge to verify that the marginal condition effects are stable across raters. Per-judge model: $\text{correct} \sim \text{condition} + (1 \mid \text{instance})$.

\begin{table}[H]
\centering\small
\adjustbox{max width=\columnwidth}{%
\begin{tabular}{lcccc}
\toprule
\textbf{Judge} & Intercept & Real NLE & Placebo NLE & Notes \\
\midrule
DeepSeek-R1 & 7.94$\times 10^{5}$ & 0.377 & 0.131 & singular/near-separation \\
GPT-4o      & 0.443 & 1.502 & 1.228 & stable \\
\bottomrule
\end{tabular}%
}
\caption[E2 judge-specific ORs.]{E2 judge-specific ORs. DeepSeek-R1's per-judge fit is near-singular (extreme intercept) but directionally consistent with GPT-4o: neither judge shows a significant NLE effect on accuracy.}
\label{tab:app-e2-judge}
\end{table}

\paragraph{Generator and same-family.} Fitted on NLE rows ($N = 480$, 2 conditions):

\begin{table}[H]
\centering\small
\adjustbox{max width=\columnwidth}{%
\begin{tabular}{lrrr}
\toprule
\textbf{Effect} & \textbf{Estimate} & \textbf{SE} & $p$ \\
\midrule
Intercept              & $-$0.060 & 0.397 & .880 \\
Placebo NLE            & $-$0.163 & 0.242 & .502 \\
Generator: GPT-4o      & $+$0.358 & 0.243 & .141 \\
Judge: GPT-4o          & $-$0.810 & 0.248 & .001 \\
\midrule
Same-family (OR $= 1.528$) & \multicolumn{2}{c}{$p = .082$ (marginal)} & \\
\bottomrule
\end{tabular}%
}
\caption[E2 generator and same-family tests.]{E2 generator and same-family tests. No generator effect. Same-family bias is marginal ($p = .082$) but does not survive adjustment for multiple tests.}
\label{tab:app-e2-gen}
\end{table}

\paragraph{Random slopes.} Singular fit; intercept-only model retained.

\paragraph{Derangement caveat.} Placebo NLEs were constructed using a single random derangement of the 60 instances (each instance receives an NLE generated for a different instance, with the constraint that no instance receives its own). A different derangement would produce slightly different placebo pairings; results are expected to be stable given the large $N$ but the dependence on one specific assignment is acknowledged.

\subsection{Experiment 3: Counterfactual Simulatability}
\label{app:p2-e3}

\paragraph*{Descriptives}

\begin{table}[H]
\centering\small
\adjustbox{max width=\columnwidth}{%
\begin{tabular}{lcccc}
\toprule
\textbf{Condition} & \textbf{$N$} & \textbf{Accuracy} & \textbf{Mean conf.} & \textbf{Median conf.} \\
\midrule
NLE + SHAP (E+X) & 240 & .579 & 4.00 & 4 \\
SHAP only (X)   & 120 & .533 & 3.97 & 4 \\
\bottomrule
\end{tabular}%
}
\caption[E3 descriptive statistics by condition.]{E3 descriptives. True-direction distribution across the 60 test instances: higher 28 (168 judgments), lower 24 (144), similar 8 (48). By judge: E+X/DeepSeek .617/4.05, E+X/GPT-4o .542/3.95; X/DeepSeek .600/3.95, X/GPT-4o .467/3.98.}
\label{tab:app-e3-desc}
\end{table}

\paragraph*{GLMM Fixed Effects (Accuracy)}

\begin{table}[H]
\centering\small
\adjustbox{max width=\columnwidth}{%
\begin{tabular}{lrrrr}
\toprule
\textbf{Effect} & \textbf{Estimate} & \textbf{SE} & \textbf{OR} & \textbf{$p$} \\
\midrule
Intercept              & $+$0.829 & 0.357 & 2.292 & .020 \\
Condition X            & $-$0.116 & 0.401 & 0.891 & .772 \\
Judge: GPT-4o          & $-$0.514 & 0.327 & 0.599 & .116 \\
X $\times$ Judge       & $-$0.391 & 0.563 & 0.677 & .488 \\
\midrule
Random intercept SD    & \multicolumn{4}{c}{2.006 (variance 4.025; 60 groups)} \\
Fit                    & \multicolumn{4}{c}{AIC $=$ 419.2; BIC $=$ 438.6; $\log L = -204.6$} \\
\bottomrule
\end{tabular}%
}
\caption[E3 GLMM fixed effects.]{E3 GLMM fixed effects. Reference: E+X, DeepSeek-R1. $N = 360$, 60 instance groups. No interaction significant.}
\label{tab:app-e3-glmm}
\end{table}

\paragraph*{Estimated Marginal Means (Accuracy)}

\begin{table}[H]
\centering\small
\adjustbox{max width=\columnwidth}{%
\begin{tabular}{lccc}
\toprule
\textbf{Condition} & \textbf{Prob.} & \textbf{SE} & \textbf{95\% CI} \\
\midrule
\multicolumn{4}{l}{\textit{Marginal:}} \\
NLE + SHAP (E+X) & .639 & .074 & [.480, .772] \\
SHAP only (X)    & .565 & .085 & [.397, .717] \\
\midrule
\multicolumn{4}{l}{\textit{DeepSeek-R1:}} \\
E+X & .696 & .075 & [.530, .821] \\
X   & .671 & .090 & [.477, .820] \\
\midrule
\multicolumn{4}{l}{\textit{GPT-4o:}} \\
E+X & .578 & .083 & [.412, .729] \\
X   & .452 & .093 & [.281, .633] \\
\bottomrule
\end{tabular}%
}
\caption[E3 estimated marginal means.]{E3 EMMs on the probability scale. NLE point estimates are higher for both judges but wide CIs span the SHAP-only values.}
\label{tab:app-e3-emm}
\end{table}

\paragraph*{Confusion Matrices}

\begin{table}[H]
\centering\small
\adjustbox{max width=\columnwidth}{%
\begin{tabular}{lccc|c}
\toprule
\textbf{Predicted} & \textbf{True: higher} & \textbf{lower} & \textbf{similar} & \textbf{Row total} \\
\midrule
\multicolumn{5}{l}{\textit{SHAP only (X):}} \\
higher  & 31 & 10 & 9 & 50 \\
lower   & 19 & 30 & 4 & 53 \\
similar & 6  & 8  & 3 & 17 \\
\midrule
\multicolumn{5}{l}{\textit{NLE + SHAP (E+X):}} \\
higher  & 70 & 18 & 17 & 105 \\
lower   & 32 & 61 & 7  & 100 \\
similar & 10 & 17 & 8  & 35 \\
\bottomrule
\end{tabular}%
}
\caption[E3 confusion matrices.]{E3 confusion matrices for both conditions (rows: predicted; columns: true). Per-class recall (diagonal / column sum): see next table.}
\label{tab:app-e3-confusion}
\end{table}

\begin{table}[H]
\centering\small
\adjustbox{max width=\columnwidth}{%
\begin{tabular}{lccc}
\toprule
\textbf{Condition} & \textbf{Higher recall} & \textbf{Lower recall} & \textbf{Similar recall} \\
\midrule
NLE + SHAP (E+X) & .625 (70/112) & .635 (61/96) & \textbf{.250 (8/32)} \\
SHAP only (X)    & .554 (31/56)  & .625 (30/48) & \textbf{.188 (3/16)} \\
\bottomrule
\end{tabular}%
}
\caption[E3 per-class recall.]{E3 per-class recall. The ``similar'' class (small perturbation, predicted output barely changes) is substantially harder for both conditions; neither NLE nor SHAP-only reaches even 25\% recall.}
\label{tab:app-e3-recall}
\end{table}

\paragraph*{CLMM (Confidence)}

\begin{table}[H]
\centering\small
\adjustbox{max width=\columnwidth}{%
\begin{tabular}{lrrrr}
\toprule
\textbf{Effect} & \textbf{Estimate} & \textbf{SE} & \textbf{OR} & \textbf{$p$} \\
\midrule
Condition X            & $-$0.528 & 0.365 & 0.590 & .148 \\
Judge: GPT-4o          & $-$0.480 & 0.293 & 0.619 & .101 \\
X $\times$ Judge       & $+$0.678 & 0.506 & 1.970 & .180 \\
\midrule
Threshold $\theta_{3|4}$ & $-$2.508 & 0.334 & & \\
Threshold $\theta_{4|5}$ & $+$2.006 & 0.319 & & \\
Random intercept SD     & \multicolumn{4}{c}{1.466 (variance 2.149; 60 groups)} \\
\bottomrule
\end{tabular}%
}
\caption[E3 CLMM fixed effects.]{E3 CLMM on confidence. Confidence is near ceiling in both conditions; no coefficient reaches significance.}
\label{tab:app-e3-clmm}
\end{table}

\paragraph*{Calibration}

\begin{table}[H]
\centering\small
\adjustbox{max width=\columnwidth}{%
\begin{tabular}{lrrrr}
\toprule
\textbf{Effect} & \textbf{Estimate} & \textbf{SE} & $z$ & \textbf{$p$} \\
\midrule
Intercept                      & $-$2.600 & 1.505 & $-$1.73 & .084 \\
\textbf{confidence\_num}       & $\mathbf{+0.875}$ & \textbf{0.367} & $\mathbf{+2.39}$ & $\mathbf{.017}$ \\
Condition X                    & $+$3.368 & 2.210 & $+$1.52 & .128 \\
Judge: GPT-4o                  & $-$0.604 & 0.284 & $-$2.12 & .033 \\
conf $\times$ X                & $-$0.932 & 0.559 & $-$1.67 & .095 \\
\midrule
Random intercept SD            & \multicolumn{4}{c}{1.928} \\
\bottomrule
\end{tabular}%
}
\caption[E3 calibration model coefficients.]{E3 calibration model. The main \texttt{confidence\_num} coefficient (the E+X calibration slope) is positive and significant -- higher confidence genuinely predicts accuracy under NLE. The conf~$\times$~X interaction is marginal ($p = .095$), suggesting the SHAP-only condition has a weaker slope (the $D$-values corroborate this).}
\label{tab:app-e3-cal}
\end{table}

\begin{table}[H]
\centering\small
\adjustbox{max width=\columnwidth}{%
\begin{tabular}{lcc}
\toprule
\textbf{Condition} & \textbf{Somers' $D$} & $N$ \\
\midrule
NLE + SHAP (E+X) & $+$0.259 & 240 \\
SHAP only (X)    & $+$0.089 & 120 \\
\bottomrule
\end{tabular}%
}
\caption[E3 Somers' $D$ by condition.]{E3 Somers' $D$. Both conditions show positive calibration; the NLE condition is substantially more informative per unit of confidence.}
\label{tab:app-e3-somers}
\end{table}

\begin{table}[H]
\centering\small
\adjustbox{max width=\columnwidth}{%
\begin{tabular}{llcc}
\toprule
\textbf{Condition} & \textbf{Confidence level} & \textbf{Accuracy} & $n$ \\
\midrule
NLE + SHAP (E+X) & 3 & .237 & 38 \\
                 & 4 & .628 & 164 \\
                 & 5 & .711 & 38 \\
SHAP only (X)    & 3 & .409 & 22 \\
                 & 4 & .563 & 80 \\
                 & 5 & .556 & 18 \\
\bottomrule
\end{tabular}%
}
\caption[E3 empirical calibration.]{E3 empirical calibration. The E+X curve rises monotonically (24\% at conf~3 $\to$ 71\% at conf~5); the X-only curve is flat from conf~4 to conf~5.}
\label{tab:app-e3-calcurve}
\end{table}

\begin{table}[H]
\centering\small
\adjustbox{max width=\columnwidth}{%
\begin{tabular}{lcccc}
\toprule
\textbf{Condition} & \textbf{Conf.\ | incorrect} & $n$ & \textbf{Conf.\ | correct} & $n$ \\
\midrule
NLE + SHAP (E+X) & 3.822 & 101 & 4.129 & 139 \\
SHAP only (X)    & 3.911 &  56 & 4.016 &  64 \\
\bottomrule
\end{tabular}%
}
\caption[E3 conditional confidence.]{E3 mean confidence split by correctness. In both conditions, correct judgments carry higher confidence than incorrect -- the positive calibration story.}
\label{tab:app-e3-condconf}
\end{table}

\paragraph*{Diagnostics}

Convergence: OK. Singularity: OK. DHARMa uniformity $p = .927$. Dispersion $p = .928$. No violations.

\paragraph*{Bayesian ROPE}

\begin{table}[H]
\centering\small
\adjustbox{max width=\columnwidth}{%
\begin{tabular}{lccc}
\toprule
\textbf{Parameter} & \textbf{Post.\ OR} & \textbf{95\% CrI} & \textbf{ROPE \%} \\
\midrule
Intercept      & 2.498 & [1.216, 5.460] & -- \\
Judge: GPT-4o  & 0.525 & [0.301, 0.890] & -- \\
Condition X    & 0.732 & [0.416, 1.306] & 27.9 \\
\bottomrule
\end{tabular}%
}
\caption[E3 Bayesian ROPE analysis.]{E3 Bayesian posterior ORs. The Condition X posterior's CrI spans 1; 27.9\% of its mass is inside the ROPE, consistent with a small, non-significant negative effect on accuracy relative to E+X.}
\label{tab:app-e3-rope}
\end{table}

\paragraph*{Sensitivity}

\paragraph{Judge-specific.} Stats-plan item: refit the primary GLMM separately per judge to verify that the condition effect is stable across raters. Per-judge model: $\text{correct} \sim \text{condition} + (1 \mid \text{instance})$.

\begin{table}[H]
\centering\small
\adjustbox{max width=\columnwidth}{%
\begin{tabular}{lcc}
\toprule
\textbf{Judge} & \textbf{Condition X (OR)} & \textbf{Significant?} \\
\midrule
DeepSeek-R1 & 0.84 & no \\
GPT-4o      & 0.47 & no \\
\bottomrule
\end{tabular}%
}
\caption[E3 judge-specific ORs.]{E3 judge-specific condition ORs. Neither judge's effect is individually significant; both directions agree (NLE slightly reduces the odds of a correct direction prediction relative to SHAP-only).}
\label{tab:app-e3-judge}
\end{table}

\paragraph{Generator and same-family.} Fitted on NLE rows ($N = 240$):

\begin{table}[H]
\centering\small
\adjustbox{max width=\columnwidth}{%
\begin{tabular}{lrrr}
\toprule
\textbf{Effect} & \textbf{Estimate} & \textbf{SE} & $p$ \\
\midrule
Intercept                  & $+$0.858 & 0.399 & .032 \\
Generator: GPT-4o          & $-$0.058 & 0.324 & .858 \\
Judge: GPT-4o              & $-$0.520 & 0.327 & .112 \\
\midrule
Same-family (OR $= 1.516$) & \multicolumn{2}{c}{$p = .232$} & \\
\bottomrule
\end{tabular}%
}
\caption[E3 generator and same-family tests.]{E3 generator and same-family tests. No generator effect; no significant same-family bias.}
\label{tab:app-e3-gen}
\end{table}

\paragraph{Random slopes.} Singular fit; intercept-only model retained.

\subsection{Experiment 4: Mental Model Transfer}
\label{app:p2-e4}

\paragraph*{Descriptives}

\begin{table}[H]
\centering\small
\adjustbox{max width=\columnwidth}{%
\begin{tabular}{lcccc}
\toprule
\textbf{Condition} & \textbf{$N$} & \textbf{Accuracy} & \textbf{Mean conf.} & \textbf{Median conf.} \\
\midrule
Baseline (no NLE training) & 110 & .364 & 3.57 & 4 \\
NLE training               & 220 & .405 & 3.57 & 4 \\
\bottomrule
\end{tabular}%
}
\caption[E4 descriptive statistics by condition.]{E4 descriptives across the 55 sliding-window test positions. True-bucket distribution: small 72, medium 150, large 90, very-large 18 judgments. By judge: Baseline/DeepSeek .309/3.40, Baseline/GPT-4o .418/3.75; E/DeepSeek .400/3.40, E/GPT-4o .409/3.75.}
\label{tab:app-e4-desc}
\end{table}

\paragraph*{GLMM Fixed Effects (Accuracy)}

\begin{table}[H]
\centering\small
\adjustbox{max width=\columnwidth}{%
\begin{tabular}{lrrrr}
\toprule
\textbf{Effect} & \textbf{Estimate} & \textbf{SE} & \textbf{OR} & \textbf{$p$} \\
\midrule
Intercept              & $-$3.387 & 1.257 & 0.034 & .007 \\
Condition E            & $+$1.489 & 0.708 & 4.433 & .035 \\
Judge: GPT-4o          & $+$1.763 & 0.804 & 5.831 & .028 \\
E $\times$ Judge       & $-$1.625 & 0.954 & 0.197 & .089 \\
\midrule
Random intercept SD    & \multicolumn{4}{c}{5.359 (variance 28.72; 55 groups)} \\
Fit                    & \multicolumn{4}{c}{AIC $=$ 249.5; BIC $=$ 268.5; $\log L = -119.7$} \\
\bottomrule
\end{tabular}%
}
\caption[E4 GLMM fixed effects.]{E4 GLMM fixed effects. Reference: Baseline, DeepSeek-R1. Omnibus LRT: $\chi^2(2) = 4.86$, $p = .089$. $N = 330$, 55 instance groups. The extreme random-intercept variance reflects large between-instance difficulty variation specific to this experiment's sliding-window setup. The judge~$\times$~condition interaction is marginal ($p = .089$), indicating opposing patterns across judges.}
\label{tab:app-e4-glmm}
\end{table}

\paragraph*{Estimated Marginal Means (Accuracy)}

\begin{table}[H]
\centering\small
\adjustbox{max width=\columnwidth}{%
\begin{tabular}{lcc}
\toprule
\textbf{Condition} & \textbf{Marginal prob.} & \textbf{DeepSeek-R1 / GPT-4o} \\
\midrule
Baseline     & .076 & .033 / .165 \\
NLE training & .138 & .130 / .147 \\
\bottomrule
\end{tabular}%
}
\caption[E4 estimated marginal means.]{E4 EMMs on the probability scale. Standard errors and confidence intervals are not reported because the extreme instance-level random intercept (SD $= 5.36$, variance $= 28.72$) makes them uninformative at these near-floor probability levels. The judge-specific EMMs reveal the key asymmetry: DeepSeek-R1's accuracy improves from 3.3\% to 13.0\% with NLE training; GPT-4o's slightly decreases (16.5\% $\to$ 14.7\%).}
\label{tab:app-e4-emm}
\end{table}

\paragraph*{CLMM (Confidence)}

\begin{table}[H]
\centering\small
\adjustbox{max width=\columnwidth}{%
\begin{tabular}{lrrrr}
\toprule
\textbf{Effect} & \textbf{Estimate} & \textbf{SE} & \textbf{OR} & \textbf{$p$} \\
\midrule
Condition E            & $-$0.019 & 0.366 & 0.981 & .958 \\
\textbf{Judge: GPT-4o} & $\mathbf{+1.657}$ & \textbf{0.451} & \textbf{5.243} & $\mathbf{.0002}$ \\
E $\times$ Judge       & $+$0.018 & 0.535 & 1.018 & .973 \\
\midrule
Threshold $\theta_{2|3}$ & $-$5.717 & 1.054 & & \\
Threshold $\theta_{3|4}$ & $+$0.466 & 0.329 & & \\
Threshold $\theta_{4|5}$ & $+$6.246 & 0.716 & & \\
Random intercept SD      & \multicolumn{4}{c}{0.962 (55 groups)} \\
\bottomrule
\end{tabular}%
}
\caption[E4 CLMM fixed effects.]{E4 CLMM on confidence. The condition effect is virtually zero ($p = .96$); the only meaningful coefficient is judge (GPT-4o reports systematically higher confidence). Training-phase NLE exposure has no residual effect on confidence reported at test time.}
\label{tab:app-e4-clmm}
\end{table}

\paragraph*{Calibration}

\begin{table}[H]
\centering\small
\adjustbox{max width=\columnwidth}{%
\begin{tabular}{lrrrr}
\toprule
\textbf{Effect} & \textbf{Estimate} & \textbf{SE} & $z$ & \textbf{$p$} \\
\midrule
Intercept                    & $+$2.973 & 3.383 & $+$0.88 & .380 \\
\textbf{confidence\_num}     & $\mathbf{-1.964}$ & \textbf{0.930} & $\mathbf{-2.11}$ & $\mathbf{.035}$ \\
Condition E                  & $+$3.372 & 3.645 & $+$0.93 & .355 \\
\textbf{Judge: GPT-4o}       & $\mathbf{+1.799}$ & \textbf{0.636} & $\mathbf{+2.83}$ & $\mathbf{.005}$ \\
conf $\times$ E              & $-$0.755 & 1.033 & $-$0.73 & .465 \\
\midrule
Random intercept SD          & \multicolumn{4}{c}{6.316} \\
\bottomrule
\end{tabular}%
}
\caption[E4 calibration model coefficients.]{E4 calibration model. The significant negative main \texttt{confidence\_num} slope -- in both conditions per Somers' $D$ below -- indicates that higher stated confidence actually predicts \emph{lower} accuracy. Training-phase NLE exposure does not change this (interaction $p = .46$).}
\label{tab:app-e4-cal}
\end{table}

\begin{table}[H]
\centering\small
\adjustbox{max width=\columnwidth}{%
\begin{tabular}{lcc}
\toprule
\textbf{Condition} & \textbf{Somers' $D$} & $N$ \\
\midrule
Baseline     & $-$0.093 & 110 \\
NLE training & $-$0.143 & 220 \\
\bottomrule
\end{tabular}%
}
\caption[E4 Somers' $D$ by condition.]{E4 Somers' $D$. Both negative -- inverted calibration: higher stated confidence predicts \emph{lower} accuracy in both conditions.}
\label{tab:app-e4-somers}
\end{table}

\begin{table}[H]
\centering\small
\adjustbox{max width=\columnwidth}{%
\begin{tabular}{llcc}
\toprule
\textbf{Condition} & \textbf{Confidence level} & \textbf{Accuracy} & $n$ \\
\midrule
Baseline     & 2 & 1.00 & 1 \\
             & 3 & .391 & 46 \\
             & 4 & .339 & 62 \\
             & 5 & .000 & 1 \\
NLE training & 3 & .479 & 96 \\
             & 4 & .352 & 122 \\
             & 5 & .000 & 2 \\
\bottomrule
\end{tabular}%
}
\caption[E4 empirical calibration.]{E4 empirical calibration. Both conditions show declining accuracy from conf~3 to conf~4 -- the inversion captured by negative Somers' $D$.}
\label{tab:app-e4-calcurve}
\end{table}

\begin{table}[H]
\centering\small
\adjustbox{max width=\columnwidth}{%
\begin{tabular}{lcccc}
\toprule
\textbf{Condition} & \textbf{Conf.\ | incorrect} & $n$ & \textbf{Conf.\ | correct} & $n$ \\
\midrule
Baseline     & 3.614 &  70 & 3.500 &  40 \\
NLE training & 3.634 & 131 & 3.483 &  89 \\
\bottomrule
\end{tabular}%
}
\caption[E4 conditional confidence.]{E4 mean confidence split by correctness. Incorrect judgments carry \emph{higher} confidence than correct -- again, the inversion.}
\label{tab:app-e4-condconf}
\end{table}

\paragraph*{Diagnostics}

Convergence: OK. Singularity: OK. DHARMa uniformity $p = .152$. Dispersion $p = .504$. No violations.

\paragraph*{Bayesian ROPE}

\begin{table}[H]
\centering\small
\adjustbox{max width=\columnwidth}{%
\begin{tabular}{lccc}
\toprule
\textbf{Parameter} & \textbf{Post.\ OR} & \textbf{95\% CrI} & \textbf{ROPE \%} \\
\midrule
Intercept     & 0.106 & [0.014, 0.668] & -- \\
Condition E   & 1.844 & [0.742, 4.558] & 12.8 \\
Judge: GPT-4o & 1.891 & [0.798, 4.647] & -- \\
\bottomrule
\end{tabular}%
}
\caption[E4 Bayesian ROPE analysis.]{E4 Bayesian posterior ORs. Condition E's posterior favours a positive effect but with wide uncertainty (CrI $[0.74, 4.56]$ includes 1); only 12.8\% of the posterior is inside the ROPE, so the Bayesian result is also indeterminate.}
\label{tab:app-e4-rope}
\end{table}

\paragraph*{Sensitivity}

\paragraph{Judge-specific.} Stats-plan item: refit the primary GLMM separately per judge to verify that the training-phase NLE effect is stable across raters. Per-judge model: $\text{correct} \sim \text{condition} + (1 \mid \text{instance})$.

\begin{table}[H]
\centering\small
\adjustbox{max width=\columnwidth}{%
\begin{tabular}{lccl}
\toprule
\textbf{Judge} & \textbf{Condition E (OR)} & \textbf{Significant?} & \textbf{Notes} \\
\midrule
DeepSeek-R1 & 17.2 & unreliable & near-separation (baseline acc.\ $\approx 0$) \\
GPT-4o      & 0.63 & no           & -- \\
\bottomrule
\end{tabular}%
}
\caption[E4 judge-specific ORs.]{E4 judge-specific condition ORs. The two judges disagree in direction: DeepSeek-R1's extreme estimate reflects near-separation (its Baseline accuracy is close to zero), while GPT-4o's null is stable. This asymmetry drives the marginal $p = .089$ judge~$\times$~condition interaction in the primary model.}
\label{tab:app-e4-judge}
\end{table}

\paragraph{Generator and same-family.} Fitted on NLE rows ($N = 220$):

\begin{table}[H]
\centering\small
\adjustbox{max width=\columnwidth}{%
\begin{tabular}{lrrr}
\toprule
\textbf{Effect} & \textbf{Estimate} & \textbf{SE} & $p$ \\
\midrule
Intercept                  & $-$1.638 & 1.038 & .114 \\
Generator: GPT-4o          & $-$0.120 & 0.489 & .806 \\
Judge: GPT-4o              & $+$0.120 & 0.489 & .806 \\
\midrule
Same-family (OR $= 0.887$) & \multicolumn{2}{c}{$p = .806$} & \\
\bottomrule
\end{tabular}%
}
\caption[E4 generator and same-family tests.]{E4 generator and same-family tests. Neither effect is significant.}
\label{tab:app-e4-gen}
\end{table}

\paragraph{Random slopes.} Singular fit; intercept-only model retained.

\paragraph{Trial-order covariate (learning-over-time check).} Stats-plan item: test whether judges improve with more experience.

\begin{table}[H]
\centering\small
\adjustbox{max width=\columnwidth}{%
\begin{tabular}{lrrr}
\toprule
\textbf{Effect} & \textbf{Estimate} & \textbf{SE} & $p$ \\
\midrule
scale(trial\_order)                      & $-$0.286 & 0.919 & .756 \\
Condition E $\times$ scale(trial\_order) & $-$0.842 & 0.474 & .076 \\
\bottomrule
\end{tabular}%
}
\caption[E4 trial-order covariate.]{E4 trial-order test. No main effect of position in the sequence; the marginal interaction ($p = .076$) hints at differential learning across trials by condition, though the effect is not robust.}
\label{tab:app-e4-trial}
\end{table}

\paragraph{Sliding-window caveat.} The E4 design uses a sliding window with 80\% overlap between adjacent training sets; this induces temporal autocorrelation in the judgments beyond what the instance random intercept absorbs. Inferential claims for this experiment carry this caveat.

\subsection{Experiment 5: Selective Reliance}
\label{app:p2-e5}

\paragraph*{GLMM Fixed Effects (Accuracy)}

\paragraph*{Descriptives}

\begin{table}[H]
\centering\small
\adjustbox{max width=\columnwidth}{%
\begin{tabular}{llccc}
\toprule
\textbf{Data type} & \textbf{NLE} & \textbf{$N$} & \textbf{Accuracy} & \textbf{Mean conf.} \\
\midrule
Baseline & No  & 120 & .950 & 3.98 \\
Baseline & Yes & 240 & .983 & 3.99 \\
OOD      & No  & 120 & .300 & 3.82 \\
OOD      & Yes & 240 & .150 & 3.89 \\
\bottomrule
\end{tabular}%
}
\caption[E5 descriptive statistics.]{E5 descriptives. By judge: Baseline/No NLE DeepSeek .900, GPT-4o 1.000; Baseline/NLE DeepSeek .967, GPT-4o 1.000; OOD/No NLE DeepSeek .550, GPT-4o .050; OOD/NLE DeepSeek .292, GPT-4o .008. The OOD × judge asymmetry is extreme: GPT-4o almost never flags OOD predictions as unreliable.}
\label{tab:app-e5-desc}
\end{table}

\paragraph*{GLMM Fixed Effects (Accuracy)}

\begin{table}[H]
\centering\small
\adjustbox{max width=\columnwidth}{%
\begin{tabular}{lrrrr}
\toprule
\textbf{Effect} & \textbf{Estimate} & \textbf{SE} & \textbf{OR} & \textbf{$p$} \\
\midrule
Intercept               & $+$4.755 & 0.580 & 116.2 & $<$.001 \\
OOD                     & $-$4.865 & 0.580 & 0.008 & $<$.001 \\
Has NLE                 & $+$1.189 & 0.672 & 3.283 & .077 \\
Judge: GPT-4o           & $-$2.123 & 0.344 & 0.120 & $<$.001 \\
\textbf{OOD $\times$ NLE} & $\mathbf{-2.330}$ & \textbf{0.744} & \textbf{0.097} & $\mathbf{.002}$ \\
\midrule
Random intercept SD      & \multicolumn{4}{c}{0.872 (variance 0.760; 60 groups, boundary-singular)} \\
Fit                      & \multicolumn{4}{c}{AIC $=$ 394.6; BIC $=$ 422.1; $\log L = -191.3$} \\
\bottomrule
\end{tabular}%
}
\caption[E5 GLMM fixed effects.]{E5 GLMM fixed effects. Reference: Baseline, No NLE, DeepSeek-R1. Two-way interaction LRT: $\chi^2(1) = 9.84$, $p = .002$. Three-way with judge: $\chi^2(3) = 58.78$, $p = 1.07 \times 10^{-12}$ (full-model AIC $=$ 341.8 vs two-way AIC $=$ 394.6); judge asymmetry is pronounced. $N = 720$.}
\label{tab:app-e5-glmm}
\end{table}

\paragraph*{Simple Effects}

\begin{table}[H]
\centering\small
\adjustbox{max width=\columnwidth}{%
\begin{tabular}{lccc}
\toprule
\textbf{Simple effect} & \textbf{OR} & \textbf{Interpretation} & $p$ \\
\midrule
NLE within Baseline       & 0.31 & NLE marginally helps & .077 \\
\textbf{NLE within OOD}    & $\mathbf{3.13}$ & \textbf{NLE significantly hurts} & $\mathbf{.0003}$ \\
Poisoning within No NLE   & 130 & extreme drop & $<$.001 \\
\textbf{Poisoning within NLE} & $\mathbf{1333}$ & \textbf{drop is bigger when NLE is present} & $<$.001 \\
\bottomrule
\end{tabular}%
}
\caption[E5 simple effects.]{E5 simple-effects decomposition of the significant poisoning~$\times$~NLE interaction. The reversed direction of the NLE effect between Baseline and OOD -- help to harm -- is the interaction.}
\label{tab:app-e5-simple}
\end{table}

\paragraph*{CLMM (Confidence)}

\begin{table}[H]
\centering\small
\adjustbox{max width=\columnwidth}{%
\begin{tabular}{lrrrr}
\toprule
\textbf{Effect} & \textbf{Estimate} & \textbf{SE} & \textbf{OR} & \textbf{$p$} \\
\midrule
OOD                 & $-$2.241 & 0.591 & 0.106 & .0001 \\
Has NLE             & $+$0.142 & 0.624 & 1.153 & .820 \\
\textbf{Judge: GPT-4o} & $\mathbf{+3.593}$ & \textbf{0.727} & \textbf{36.3} & $\mathbf{<.001}$ \\
OOD $\times$ NLE    & $+$0.514 & 0.713 & 1.672 & .470 \\
\midrule
Threshold $\theta_{2|3}$ & $-$6.097 & 0.776 & & \\
Threshold $\theta_{3|4}$ & $-$2.945 & 0.528 & & \\
Threshold $\theta_{4|5}$ & $+$7.655 & 0.968 & & \\
Random intercept SD      & \multicolumn{4}{c}{$\approx 0$ (confidence does not vary by instance)} \\
\bottomrule
\end{tabular}%
}
\caption[E5 CLMM fixed effects.]{E5 CLMM on confidence. Poisoning decreases confidence slightly (the OOD prediction is rated slightly less confidently), but the NLE main effect and interaction on confidence are non-significant: confidence is essentially uniform (3.82--3.99) across all four conditions. The poisoning~$\times$~NLE interaction LRT on confidence is $\chi^2(1) = 0.52$, $p = .47$.}
\label{tab:app-e5-clmm}
\end{table}

\paragraph*{Calibration}

\begin{table}[H]
\centering\small
\adjustbox{max width=\columnwidth}{%
\begin{tabular}{lrrrr}
\toprule
\textbf{Effect} & \textbf{Estimate} & \textbf{SE} & $z$ & \textbf{$p$} \\
\midrule
Intercept                     & $-$10.891 & 5.544 & $-$1.96 & .049 \\
\textbf{confidence\_num}      & $\mathbf{+3.889}$ & \textbf{1.444} & $\mathbf{+2.69}$ & $\mathbf{.007}$ \\
Baseline, NLE                 & $-$10.978 & 7.310 & $-$1.50 & .133 \\
\textbf{OOD, No NLE}          & $\mathbf{+18.447}$ & \textbf{6.012} & $\mathbf{+3.07}$ & $\mathbf{.002}$ \\
\textbf{OOD, NLE}             & $\mathbf{+16.810}$ & \textbf{5.831} & $\mathbf{+2.88}$ & $\mathbf{.004}$ \\
Judge: GPT-4o                 & $-$1.928 & 0.403 & $-$4.79 & $<$.001 \\
conf $\times$ Baseline, NLE   & $+$3.339 & 1.968 & $+$1.70 & .090 \\
\textbf{conf $\times$ OOD, No NLE} & $\mathbf{-5.931}$ & \textbf{1.560} & $\mathbf{-3.80}$ & $\mathbf{.0001}$ \\
\textbf{conf $\times$ OOD, NLE} & $\mathbf{-5.778}$ & \textbf{1.516} & $\mathbf{-3.81}$ & $\mathbf{.0001}$ \\
\midrule
Random intercept SD           & \multicolumn{4}{c}{0.68} \\
\bottomrule
\end{tabular}%
}
\caption[E5 calibration model coefficients.]{E5 calibration model. In Baseline, higher confidence predicts correctness (positive main slope). In OOD, both conf~$\times$~OOD interactions are large and negative -- confidence is \emph{inversely} related to accuracy when the input is poisoned. NLE presence does not change the calibration slope in OOD (both OOD conditions have near-identical interaction coefficients).}
\label{tab:app-e5-cal}
\end{table}

\begin{table}[H]
\centering\small
\adjustbox{max width=\columnwidth}{%
\begin{tabular}{lcc}
\toprule
\textbf{Condition} & \textbf{Somers' $D$} & $N$ \\
\midrule
Baseline, No NLE & $+$0.156 & 120 \\
Baseline, NLE    & $+$0.744 & 240 \\
OOD, No NLE      & $-$0.439 & 120 \\
OOD, NLE         & $-$0.369 & 240 \\
\bottomrule
\end{tabular}%
}
\caption[E5 Somers' $D$ by condition.]{E5 Somers' $D$. Baseline conditions show positive calibration; OOD conditions show inverted calibration -- confidence predicts being \emph{wrong}. NLE presence does not change the sign.}
\label{tab:app-e5-somers}
\end{table}

\begin{table}[H]
\centering\small
\adjustbox{max width=\columnwidth}{%
\begin{tabular}{llcc}
\toprule
\textbf{Condition} & \textbf{Confidence level} & \textbf{Accuracy} & $n$ \\
\midrule
Baseline, No NLE & 3 & .667 & 3 \\
                 & 4 & .957 & 116 \\
                 & 5 & 1.00 & 1 \\
Baseline, NLE    & 3 & .400 & 5 \\
                 & 4 & .996 & 233 \\
                 & 5 & 1.00 & 2 \\
OOD, No NLE      & 2 & 1.00 & 2 \\
                 & 3 & .842 & 19 \\
                 & 4 & .173 & 98 \\
                 & 5 & 1.00 & 1 \\
OOD, NLE         & 2 & 1.00 & 1 \\
                 & 3 & .583 & 24 \\
                 & 4 & .098 & 215 \\
\bottomrule
\end{tabular}%
}
\caption[E5 empirical calibration.]{E5 empirical calibration. In OOD conditions, stated confidence~4 corresponds to only 10--17\% accuracy, while confidence~3 corresponds to 58--84\% accuracy -- strong inversion. NLE makes this slightly worse (NLE-conf-4 has 9.8\% accuracy versus No-NLE-conf-4 at 17.3\%).}
\label{tab:app-e5-calcurve}
\end{table}

\begin{table}[H]
\centering\small
\adjustbox{max width=\columnwidth}{%
\begin{tabular}{llcccc}
\toprule
\textbf{Data type} & \textbf{NLE} & \textbf{Conf.\ | incorrect} & $n$ & \textbf{Conf.\ | correct} & $n$ \\
\midrule
Baseline & No  & 3.833 &   6 & 3.991 & 114 \\
Baseline & Yes & 3.250 &   4 & 4.000 & 236 \\
OOD      & No  & 3.964 &  84 & \textbf{3.472} &  36 \\
OOD      & Yes & 3.951 & 204 & \textbf{3.556} &  36 \\
\bottomrule
\end{tabular}%
}
\caption[E5 conditional confidence.]{E5 mean confidence split by correctness. In OOD, confidence when \emph{wrong} exceeds confidence when \emph{right} -- the diagnostic fingerprint of the rationalisation mechanism.}
\label{tab:app-e5-condconf}
\end{table}

\paragraph*{Diagnostics}

Convergence: OK. Singularity: OK. DHARMa uniformity $p = .341$. Dispersion $p = .368$. No violations.

\paragraph*{Bayesian ROPE}

\begin{table}[H]
\centering\small
\adjustbox{max width=\columnwidth}{%
\begin{tabular}{lccc}
\toprule
\textbf{Parameter} & \textbf{Post.\ OR} & \textbf{95\% CrI} & \\
\midrule
Intercept               & 85.530 & [33.307, 239.558] & -- \\
OOD                     & 0.010  & [0.004, 0.026]    & -- \\
Has NLE                 & 2.704  & [0.978, 7.667]    & marginal \\
Judge: GPT-4o           & 0.149  & [0.079, 0.273]    & -- \\
\textbf{OOD $\times$ NLE} & \textbf{0.121} & \textbf{[0.038, 0.380]} & \textbf{CrI excludes 1} \\
\bottomrule
\end{tabular}%
}
\caption[E5 Bayesian posterior ORs.]{E5 Bayesian posterior ORs. The OOD~$\times$~NLE interaction is the only Part~2 effect whose Bayesian CrI excludes 1 -- positive evidence for a genuine interaction, in the direction of harm (NLE makes OOD detection worse).}
\label{tab:app-e5-rope}
\end{table}

\paragraph*{Sensitivity}

\paragraph{Judge-specific.} Stats-plan item: refit the primary GLMM separately per judge to verify that the poisoning\,$\times$\,NLE interaction is stable across raters. Per-judge model: $\text{correct} \sim \text{poisoning} \times \text{has\_nle} + (1 \mid \text{instance})$.

\begin{table}[H]
\centering\small
\adjustbox{max width=\columnwidth}{%
\begin{tabular}{lcccc}
\toprule
\textbf{Judge} & Intercept & OOD & Has NLE & OOD $\times$ NLE \\
\midrule
DeepSeek-R1 & 14.953 & 0.088 & 3.487 & 0.069 \\
GPT-4o      & 6.18$\times 10^{9}$ (sep.) & $\approx 0$ & 1.00 & 0.16 \\
\bottomrule
\end{tabular}%
}
\caption[E5 judge-specific ORs.]{E5 judge-specific ORs. GPT-4o's fit is effectively separated (GPT almost never flags OOD: 0.8--5\% accuracy), so its coefficients are unreliable in magnitude but directionally consistent with DeepSeek-R1: both judges show the same interaction direction (NLE halves OOD detection).}
\label{tab:app-e5-judge}
\end{table}

\paragraph{Generator.} Not testable -- splitting by poisoning produces too few NLE-condition rows with balanced generators after filtering.

\paragraph{Random slopes.} Singular fit; intercept-only model retained. The primary model's random-intercept variance is already near the boundary (the descriptives show Baseline accuracy essentially at ceiling), so richer random structures cannot be identified from this data.

\end{document}